%% file: vmpc.tex
\documentclass{article}

\usepackage[final]{corl_2019} 

\usepackage{times}
\usepackage[utf8]{inputenc} 
\usepackage[T1]{fontenc}    
\usepackage{url}            
\usepackage{booktabs}       
\usepackage{amsfonts}       
\usepackage{nicefrac}       
\usepackage{microtype}      
\usepackage{balance}      
\usepackage[font=footnotesize,labelfont=bf]{caption}
\usepackage{wrapfig}
\usepackage{bibunits}
\usepackage{enumitem}

\input{./macros.tex}

\title{Combining Optimal Control and Learning for \\
Visual Navigation in Novel Environments} 

 \author{Somil Bansal${}^{*1}$\quad  
 Varun Tolani${}^{*1}$\quad  
 Saurabh Gupta${}^{2}$\quad
 Jitendra Malik${}^{1,2}$\quad 
 Claire Tomlin${}^{1}$}

\begin{document}

\maketitle
\thispagestyle{plain}
\pagestyle{plain}

\begin{abstract}
Model-based control is a popular paradigm for robot navigation because it can leverage a known dynamics model to efficiently plan robust robot trajectories.
However, it is challenging to use model-based methods in settings where the environment is \apriori unknown and can only be observed partially through on-board sensors on the robot. 
In this work, we address this short-coming by coupling model-based control with learning-based perception.
The learning-based perception module produces a series of \textit{waypoints} that guide the robot to the goal via a collision-free path.
These waypoints are used by a model-based planner to generate a smooth and dynamically feasible trajectory that is executed on the physical system using feedback control.
Our experiments in simulated real-world cluttered environments and on an actual ground vehicle demonstrate that the proposed approach can reach goal locations more reliably and efficiently in novel environments as compared to purely geometric mapping-based or end-to-end learning-based alternatives.
Our approach does not rely on detailed explicit 3D maps of the environment, works well with low frame rates, and generalizes well from simulation to the real world.
Videos describing our approach and experiments are available on the project website$^3$.
\end{abstract}

\input{./sections/introduction}
\input{./sections/related_work}
\input{./sections/prob_statement}
\input{./sections/our_approach}
\input{./sections/simulations}
\input{./sections/experiments3}
\input{./sections/conclusion}
\input{./sections/acknowledgment}

\newpage
\bibliography{references}

\newpage
\input{./sections/appendix}
\end{document}

%% file: macros.tex
\usepackage{graphicx,caption,subcaption,xspace}
\usepackage{amsmath,amssymb,stmaryrd,mathtools}
\usepackage{algorithm}
\usepackage{algpseudocode}
\usepackage{acronym}
\usepackage{comment}
\usepackage{color}
\usepackage{units}
\usepackage{textcomp}

\usepackage[capitalise]{cleveref}

\DeclareCaptionLabelSeparator{periodspace}{.\quad}
\crefname{equation}{}{} 
\crefname{section}{Sec.}{Sec.}

\newcommand{\state}{z} 
\newcommand{\ctrl}{u} 
\newcommand{\pos}{p} 

\renewcommand{\time}{t}
\newcommand{\image}{I} 
\newcommand{\env}{\mathcal{E}}
\newcommand{\horizon}{H} 
\newcommand{\waypt}{\hat{w}} 

\newcommand{\metName}{LB-WayPtNav\xspace}

\providecommand{\ie}{\textit{i.e.}\xspace}
\providecommand{\eg}{\textit{e.g.}\xspace}
\providecommand{\etc}{\textit{etc.}\xspace}
\providecommand{\etal}{\textit{et al.}\xspace}
\providecommand{\vs}{\textit{vs.}\xspace}

\providecommand{\apriori}{\textit{a priori}\xspace}

\algdef{SE}[SUBALG]{Indent}{EndIndent}{}{\algorithmicend\ }%
\algtext*{Indent}
\algtext*{EndIndent}

\newcommand\blfootnote[1]{%
  \begingroup
  \renewcommand\thefootnote{}\footnote{#1}%
  \addtocounter{footnote}{-1}%
  \endgroup
}

%% file: sections/introduction.tex
\section{Introduction} \label{sec:intro}
Autonomous robot navigation
is a fundamental and well-studied problems in robotics.
However, developing a fully autonomous robot that can 
navigate in \textit{a priori} unknown environments is 
difficult due to challenges that span dynamics modeling, 
on-board perception, localization and mapping, trajectory 
generation, and optimal control.
\blfootnote{$^*$The first two authors contributed equally to this paper.}
\blfootnote{$^{1}$ University of California, Berkeley.}
\blfootnote{$^{2}$ Facebook AI Research.}
\blfootnote{$^{3}$ Project website: \href{https://vtolani95.github.io/WayPtNav/}{\textcolor{blue}{https://vtolani95.github.io/WayPtNav/}}.}

One way to approach this problem is to generate a globally-consistent
geometric map of the environment, and use it to compute a 
collision-free trajectory to the goal using optimal control and planning schemes. 
However, the real-time generation of a globally consistent map 
tends to be computationally expensive, and can be challenging 
in texture-less environments or in the presence of transparent, 
shiny objects, or strong ambient lighting~\cite{alhwarin2014ir}.
Alternative approaches employ end-to-end learning to side-step
this explicit map estimation step. However, such approaches tend 
to be extremely sample inefficient and highly specialized to the system 
they were trained on~\cite{recht2018tour}.

\begin{figure}
    \centering
	\begin{subfigure}[b]{0.35\columnwidth}
		\centering
    		\includegraphics[width=\columnwidth]{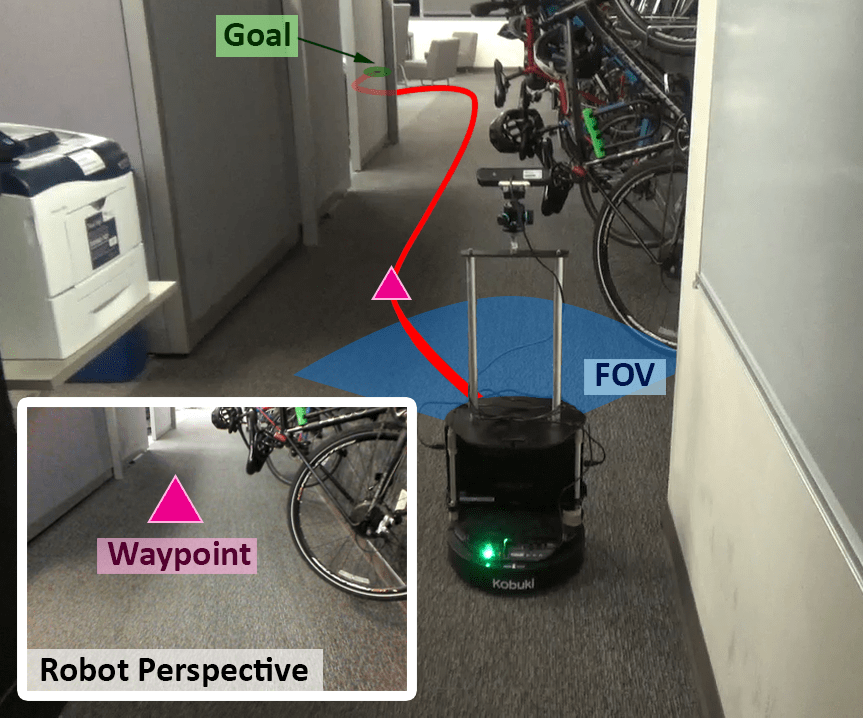}
	\end{subfigure}
	\begin{subfigure}[b]{0.55\columnwidth}
		\centering
    		\includegraphics[width=\columnwidth]{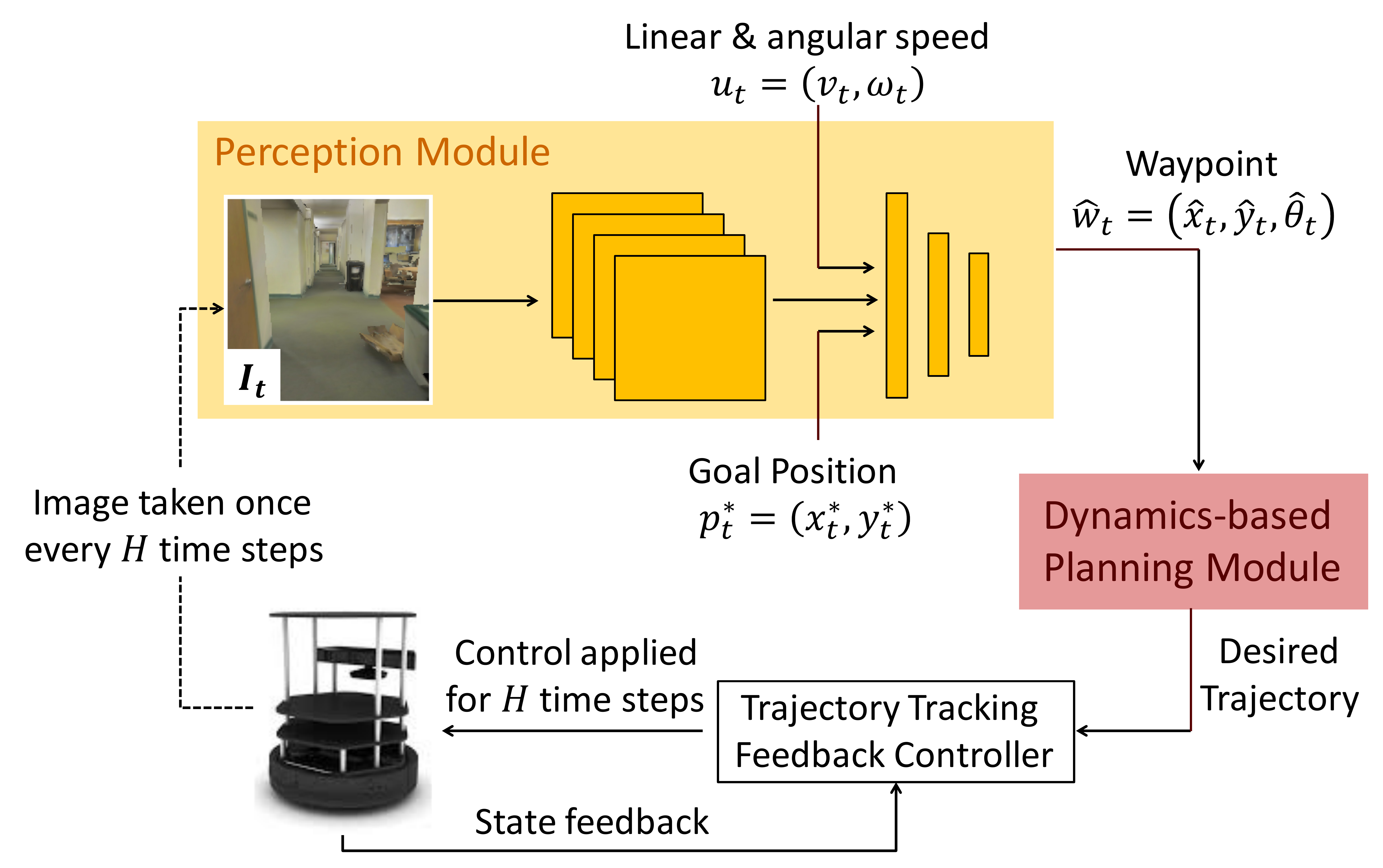}
	\end{subfigure}
	\caption{\textbf{Overview:} We consider the problem of navigation from a start position to a goal position. 
    Our approach (\metName{}) consists of a learning-based perception module and a dynamics model-based planning module. The perception module predicts a waypoint based on the current first-person RGB image observation. This waypoint is used by the model-based planning module to design a controller that smoothly regulates the system to this waypoint. This process is repeated for the next image until the robot reaches the goal.}
    \label{fig:framework}
\end{figure}

In this paper, we present a framework for autonomous, vision-based 
navigation in novel cluttered indoor environments under the assumption 
of perfect robot state measurement.
We take a factorized approach to navigation that uses \textit{learning} 
to make high-level navigation decisions in unknown environments 
and leverages \textit{optimal control} to produce smooth trajectories 
and a robust tracking controller.
In particular, we train a Convolutional Neural Network (CNN) that incrementally uses the current RGB image observations to produce a sequence of intermediate states or \textit{waypoints}. 
These waypoints are produced to guide the robot to the desired target location via a collision-free path in previously unknown environments, and are used as targets for a model-based optimal controller to generate smooth, dynamically feasible control sequences to be executed on the robot.
Our approach, \textit{\metName{}} (Learning-Based WayPoint Navigation), is summarized in Fig.~\ref{fig:framework}.

\metName benefits from the advantages of classical control and 
learning-based approaches in a way that addresses their individual limitations.
Learning can leverage statistical regularities to make 
predictions about the environment from partial views (RGB images) 
of the environment, allowing generalization to unknown environments.
Leveraging underlying dynamics and feedback-based control leads to smooth, continuous, and efficient trajectories that are naturally robust to variations in physical properties and noise in actuation, allowing us to deploy our framework directly from simulation to real-world.
Furthermore, learning now does not need to spend interaction samples to learn about the dynamics of the underlying system, and can exclusively focus on dealing with generalization to unknown environments.
To summarize, our key contributions are:
\begin{itemize}
\item an approach that combines learning and optimal control to robustly maneuver the robot in novel, cluttered environments using only a single on-board RGB camera,
\item through simulations and experiments on a mobile robot, we demonstrate that our approach is \textit{better} and more \textit{efficient} at reaching the goals, results in \textit{smoother} trajectories, as compared to End-to-End learning, and more \textit{reliable} than geometric mapping-based approaches, 
\item we demonstrate that our approach can be \textit{directly} transferred from simulation to unseen, real-world environments without any finetuning or data collection in the real-world, 
\item an optimal control method for generating optimal waypoints to support large-scale training of deep neural networks for autonomous navigation without requiring any human labeling..
\end{itemize}

%% file: sections/related_work.tex
\section{Related Work} \label{sec:related_work}
An extensive body of research studies autonomous navigation. We cannot possibly hope to summarize all these works here, but we attempt to discuss the most closely related approaches.

\textbf{Classical Robot Navigation.}
Classical robotics has made significant progress by
factorizing the problem of robot navigation into sub-problems of mapping and localization~\cite{thrun2005probabilistic,slam-survey:2015}, 
path planning~\cite{lavalle2006planning}, and trajectory tracking.
Mapping estimates the 3D structure of the world 
(using RGB / RGB-D images / LiDAR scans), which is used by a 
planner to compute paths to goal.
However, such purely geometric intermediate representations do not capture 
navigational affordances (such as: to go far away, one should step into a 
hallway, \etc).
Furthermore, mapping is challenging with just RGB observations, 
and often unreliable even with active depth sensors (such as in 
presence of shiny or thin objects, or in presence of strong 
ambient light)~\cite{alhwarin2014ir}. 
This motivates approaches that leverage object and region semantics 
during mapping and planning \cite{bowman2017probabilistic, kuipers1991robot}; 
however, such semantics are often hand-coded.
Our work is similarly motivated, but instead of using geometry-based
reasoning or hand-crafted heuristics, we employ learning to directly predict
good waypoints to convey the robot to desired target 
locations. This also side-steps the need for explicit map-building.

\textbf{End-to-End (E2E) Learning for Navigation.}
There has been a recent interest in employing end-to-end learning for training policies for goal-driven navigation \cite{zhu2016target, gupta2017cognitive, khan2017memory, kim2015deep}. The typical motivation here is to incorporate semantics and common-sense reasoning into navigation policies. While Zhu \etal \cite{zhu2016target} learn policies that work well in training environments, Gupta \etal \cite{gupta2017cognitive} and Khan \etal \cite{khan2017memory} design policies that generalize to previously unseen environments. 
Most such works abstract out dynamics and work with a set of macro-actions (going forward $x$ cm, turning $\theta^\circ$). Such ignorance of dynamics results in jerky and inefficient policies that exhibit stop-and-go behavior on a real robot. 
Several works also use end-to-end learning for navigation using 
laser scans~\cite{chiang2019learning, tai2017virtual, zeng2019end}, 
for training and combining a local planner with a higher level 
roadmap for long range navigation~\cite{gao2017intention,faust2018prm,chen2019behavioral,amini2018variational}, or visual servoing~\cite{sadeghi2019divis}.
Numerous other works have tackled navigation in synthetic game 
environments~\cite{mirowski2016learning, savinov2018semi, parisotto2017neural}, 
and largely ignore considerations of real-world deployment, such as dynamics and state estimation. 
Researchers have also employed learning to tackle locomotion 
problems~\cite{gandhi2017learning,kahn2017uncertainty,sadeghi2016cadrl,kang2019generalization}. These works learn policies for collision-avoidance, \ie how to move around in an environment without colliding. Kahn \etal~\cite{kahn2017uncertainty} use motion primitives, while Gandhi \etal \cite{gandhi2017learning}, and Sadeghi and Levine~\cite{sadeghi2016cadrl} use velocity control for locomotion. 
While all of these works implement policies for collision avoidance via 
lower level control, our work studies how policy learning itself should 
be modified for dynamically feasible low-level control for goal-driven behavior.

\textbf{Combining Optimal Control and Learning.}
A number of papers seek to combine the best of learning with optimal control for high-speed navigation~\cite{richter2018bayesian, pmlr-v78-drews17a, kaufmann2018deep, jung2018perception, loquercio2018dronet}.
Drews \etal \cite{pmlr-v78-drews17a, drews2019vision} learn a cost function from monocular images for
aggressive race-track driving via Model Predictive Control (MPC).
Kaufmann \etal \cite{kaufmann2018beauty, kaufmann2018deep} use 
learning to predict waypoints that are used with model-based 
control for drone racing. 
The focus of these works is on aggressive 
control in \textit{training race-track environments}, whereas we seek 
to learn \textit{goal-driven} policies that work well in 
\textit{completely novel, cluttered real world testing environments}.
This renders their approach for waypoint generation for learning (that does not reason about obstacles explicitly) ineffective. In contrast, waypoints generated for policy learning by our optimal control based method, are guaranteed to generate a collision free trajectory.
Muller \etal~\cite{muller2018driving} predict waypoints from 
semantically segmented images and a user provided command for 
outdoor navigation, and use a PID controller for control.
However, they do not explicitly handle obstacles and focuses primarily on lane keeping and making turns. 
Instead, we use a model-based planner to generate rich, agile, and
explicitly dynamically feasible and collision-free control behaviors, 
without requiring any user commands, and show results in 
cluttered real-world indoor environments.
In a work parallel to ours, Meng \etal~\cite{meng2019neural} combine 
Riemannian Motion Policy with learning for autonomous navigation, 
whereas we focus on dynamically feasible spline-based policies. 
Other works such as that from Levine \etal \cite{levine2016end} 
and Pan \etal \cite{pan2017agile} combine optimal control and 
end-to-end learning, by training neural network policies to mimic 
the optimal control values based on the raw images. We explicitly 
compare to such an approach in this work.

%% file: sections/prob_statement.tex
\section{Problem Setup} \label{sec:formulation}
In this work, we study the problem of autonomous navigation of a ground vehicle in previously unknown indoor environments.
We assume that odometry is perfect (\ie, the exact vehicle state is available), and that the environment is static. Dealing with imperfect odometry and dynamic environments are problems in their own right, and we defer them to future work.

We model our ground vehicle as a three-dimensional non-linear Dubins car system with dynamics:
\begin{equation} \label{eqn:NumSimpleDyn}
\dot{x} = v\cos\phi,\quad \dot{y} = v\sin\phi,\quad \dot{\phi} = \omega\,,
\end{equation}
where $\state_{\time} := (x_{\time}, y_{\time}, \phi_{\time})$ is the state of vehicle, $\pos_{\time} = (x_{\time}, y_{\time})$ is the position, $\phi_{\time}$ is the heading, $v_{\time}$ is the speed, and $\omega_{\time}$ is the turn rate at time $\time$.
The input (control) to the system is $\ctrl_{\time} := (v_{\time}, \omega_{\time})$.
The linear and angular speeds $v_{\time}$ and $\omega_{\time}$ are bounded within $[0, \bar{v}]$ and $[-\bar{\omega}, \bar{\omega}]$ respectively.
We use a discretized version of the dynamics in Eqn.~\eqref{eqn:NumSimpleDyn} for all planning purposes.
The robot is equipped with a monocular RGB camera mounted at a fixed height, oriented at a fixed pitch and pointing forwards.
The goal of this paper is to learn control policies for goal-oriented navigation tasks: the robot needs to go to a target position, $\pos^* = (x^*, y^*)$, specified in the robot's coordinate frame (\eg, $11$m forward, $5$m left), without colliding with any obstacles.
These tasks are to be performed in novel environments whose map or topology is not available to the robot.
In particular, at a given time step $\time$, the robot with state $\state_t$ receives as input an RGB image of the environment $\env$, $\image_{\time} = \image(\env, \state_{\time})$, and the target position $\pos_{\time}^* = (x_{\time}^*, y_{\time}^*)$ expressed in the current coordinate frame of the robot.
The objective is to obtain a control policy that uses these inputs to guide the robot to the target as quickly as possible.

%% file: sections/our_approach.tex
\section{Model-based Learning for Navigation} \label{sec:approach}
We use a learning-based waypoint approach to navigation (\metName{}). 
The \metName{} framework is demonstrated in Figure \ref{fig:framework} and summarized in Algorithm \ref{alg:method}.
\metName{} makes use of two submodules: perception and planning.

\subsection{\textbf{Perception Module}}
We implement the perception module using a CNN that takes as input a $224 \times 224$ pixel RGB image, $\image_{\time}$, captured from the onboard camera, the target position, $\pos_{\time}^*$, specified in the vehicle's current coordinate frame, and vehicle's current linear and angular speed, $\ctrl_{\time}$, and outputs the desired next state or a waypoint $\waypt_{\time} := (\hat{x}_{\time}, \hat{y}_{\time}, \hat{\theta}_{\time}) = \psi(I_t, u_t, p_t^*)$ (Line 5 in Algorithm \ref{alg:method}).
Intuitively, the network can use the presence of surfaces and furniture objects like floors, tables, and chairs in the scene, alongside the learned priors about their shapes to generate an estimate of the next waypoint, without explicitly building an accurate map of the environment.
This allows a more guided and efficient exploration in novel environments based on the robot's prior experience with similar scenes and objects.

\subsection{\textbf{Planning and Control Module}}
Given a waypoint $\waypt_{\time}$, and the current linear and angular speed $u_t$, the planning module uses the system dynamics in Eqn.~\eqref{eqn:NumSimpleDyn} to design a smooth trajectory, satisfying the dynamics and control constraints, from the current vehicle state to the waypoint.
In this work, we represent the $x$ and $y$ trajectories using third-order splines, whose parameters can be obtained using $\waypt_{\time}$ and $u_t$ \cite{WALAMBE2016601}.
This corresponds to solving a set of linear equations, and thus, planning can be done efficiently onboard.
Since the heading of the vehicle can be determined from the $x$ and $y$ trajectories, 
a spline-based planner ultimately provides the desired state and control trajectories, $\{z^*, u^*\}_{t:t+H} = FitSpline(\hat{w}_t, u_t)$, that the robot follows for the time horizon $[t, t+\horizon]$ to reach the waypoint $\waypt_{\time}$ (Line 6).
Since the splines are third-order, the generated speed and acceleration trajectories are smooth.
This is an important consideration for real robots, since jerky trajectories might lead to compounding sensor errors, poor tracking, or hardware damage~\cite{gonzalez2016review}.  
While we use splines in this work for computational efficiency, other model-based planning schemes can also be used for trajectory planning. 

To track the generated trajectory $\{z^*, u^*\}$, we design a LQR-based \textbf{linear feedback controller}~\cite{bender1987linear}, $\{k,~K\}_{t:t+H}~=~LQR(z^*_{t:t+H},~u^*_{t:t+H})$ (Line 7).
Here $k$ and $K$ represent the feed-forward and feedback terms respectively.
The LQR controller is obtained using the dynamics in Eqn.~\eqref{eqn:NumSimpleDyn}, linearized around the trajectory $\{z^*, u^*\}$.
LQR is a widely used feedback controller in robotic systems to make planning robust to external disturbances and mismatches between the dynamics model and the actual system~\cite{van2011lqg}.
This feedback controller allows us to deploy the proposed framework directly from simulation to a real robot (provided the real-world and simulation environments are visually similar), even though the model in Eqn.~\eqref{eqn:NumSimpleDyn} may not capture the true physics of the robot.

The control commands generated by the LQR controller are executed on the system over a time horizon of $\horizon$ seconds (Line 8), and then a new image is obtained. 
Consequently, a new waypoint and plan is generated.
This entire process is repeated until the robot reaches the goal position.
\input{./sections/algo}

\subsection{Training Details}
\metName{}'s perception module is trained via supervised learning in training
environments where the underlying map is known. 
Even though the environment map is known during training, no such assumption is made during test time and the robot relies only on an RGB image and other on-board sensors.
The knowledge of the map during training allows us to compute optimal waypoints (and trajectories) by formulating the navigation problem between randomly sampled pairs of start and goal locations as an optimal control problem, which can be solved using MPC (described in appendix in Section~\ref{appendix_sec:expert_data}).
The proposed method does not require any human labeling and can be used to generate supervision for a variety of ground and aerial vehicles.
Given first-person images and relative goal coordinates as input, the perception module is trained to predict these optimal waypoints. 

%% file: sections/algo.tex
\begin{algorithm}[t]
  \caption{Model-based Navigation via Learned Waypoint Prediction}
    \begin{algorithmic}[1]
        \Require $p^* := (x^*, y^*)$ \Comment{Goal location}\;
        \For{$t = 0$ to $T$}
            \State $z_t := (x_t, y_t, \phi_t)$;~~ $u_t := (v_t, \omega_t)$ \Comment{Measured robot pose, and linear and angular speed}
            \State \textbf{Every} $H$ steps \textbf{do} 
            \Comment{Replan every $H$ steps}
            \Indent
                \State $p^*_t := (x_t^{*}, y_t^{*})$ \Comment{Goal location in the robot's coordinate frame}
                \State $\hat{w}_t = \psi(I_t, u_t, p_t^*)$
                \Comment{Predict next waypoint}
                \State $\{z^*, u^*\}_{t:t+H} = FitSpline(\hat{w}_t, u_t)$
                \Comment{Plan spline-based smooth trajectory}
                \State $\{k, K\}_{t:t+H} = LQR(z^*_{t:t+H}, u^*_{t:t+H})$\;
                \Comment{Tracking controller}
            \EndIndent
            \State $u_{t+1} = K_{t}(\state_t-\state_{t}^{*}) + k_{t}$
            \Comment{Apply control}
        \EndFor
\end{algorithmic}
\label{alg:method}
\end{algorithm}

%% file: sections/simulations.tex
\section{Simulation Experiments} \label{sec:sims}
\metName{} is aimed at combining classical optimal control with 
learning for interpreting images. In this section, we present
experiments in simulation, and compare to representative
methods that only use E2E learning (by ignoring all knowledge
about the known system), and that only use geometric mapping and 
path planning (and no learning).

\textbf{Simulation Setup:} Our simulation experiments are conducted in 
environments derived from scans of 
real world buildings (from the Stanford large-scale 
3D Indoor Spaces dataset~\cite{armeni_cvpr16}).
Scans from 2 buildings were used to generate training data
to train \metName{}. 185 test episodes (start, goal position pairs) in 
a 3rd \textit{held-out} building were used for testing the different methods.
Test episodes are sampled to include scenarios such as: going around obstacles, 
going out of the room, going from one hallway to another.
Though training and test environments consists of indoor offices and labs,
their layouts and visual appearances are quite different (see 
Section \ref{sec:appendix:training_test_areas} for some images).

\textbf{Implementation Details:} We used a pre-trained 
ResNet-50~\cite{DBLP:journals/corr/HeZRS15} as the CNN backbone for the 
perception module, and finetuned it for waypoint prediction with MSE loss using
the Adam optimizer. More training details and the exact CNN architecture are
provided in appendix in Section~\ref{appendix_sec:arch}. 

\textbf{Comparisons:} We compare to two alternative approaches.
\textit{E2E Learning:} This approach is trained to directly output 
the velocity commands corresponding to the optimal trajectories produced by the 
spline-based planner (the same trajectories used to generate 
supervision for \metName{}, see Sec.~\ref{appendix_sec:expert_data}). This represents a purely learning
based approach that does not explicitly use any system knowledge at test time.
\textit{Geometric Mapping and Planning:} This approach represents
a learning-free, purely geometric approach. As inferring precise
geometry from RGB images is challenging, we provide ideal depth images as input
to this approach. These depth images are used to incrementally build up an occupancy 
map of the environment, that is used with the same spline-based planner 
(that was used to generate the expert supervision, see 
Sec. \ref{appendix_sec:expert_data}), to output the velocity controls.
Results reported here are with control horizon, $\horizon=1.5s$. We also 
tried $H = 0.5, 1.0$, but the results and trends were the same.

\textbf{Metrics:}
We make comparisons via the following metrics: 
success rate (if the robot reaches within $0.3m$ of the goal position without any collisions), the average time to reach the goal (for the successful episodes), and the average acceleration and jerk along the robot trajectory. The latter metrics measure execution smoothness and efficiency with respect to time and power. 
\begin{table}
\caption{\textbf{Quantitative Comparisons in Simulation:} 
Various metrics for different approaches across the test navigation tasks: success rate (higher is better), average time to reach goal, jerk and acceleration along the robot trajectory (lower is better) for successful episodes.
\metName{} conveys the robot to the goal location more often, faster, and produces considerably less jerky trajectories than E2E learning approach.
Since \metName{} only uses the current RGB image, whereas the geometric mapping and planning approach integrates information from perfect depth images, it outperforms \metName{} in simulation. However, performance is comparable when the mapping based approach only uses the current image (like \metName{}, but still depth \vs RGB).}
\label{table:success_bar}
\centering
\resizebox{1.0\linewidth}{!}{
\begin{tabular}{lccccc}
\toprule
\textbf{Agent} & \textbf{Input} & \textbf{Success (\%)} & \textbf{ Time taken (s)} & \textbf{Acceleration ($m/s^{2}$)} & \textbf{Jerk ($m/s^{3}$)} \\ \midrule
Expert & Full map &100 &      10.78 \textpm 2.64 &    0.11 \textpm 0.03 &    0.36 \textpm 0.14 \\
\midrule
\metName{} (our) & RGB & 80.65 &    11.52 \textpm 3.00 &    0.10 \textpm 0.04 &    0.39 \textpm 0.16\\
End To End & RGB & 58.06 &    19.16 \textpm 10.45 &    0.23 \textpm 0.02 &    8.07 \textpm 0.94 \\
Mapping (memoryless)  & Depth & 86.56 &     10.96 \textpm 2.74 &    0.11 \textpm 0.03 &   0.36 \textpm 0.14 \\
\midrule
Mapping  & Depth + Spatial Memory & 97.85 &  10.95 \textpm 2.75 &    0.11 \textpm 0.03 &    0.36 \textpm 0.14 \\
\bottomrule
\end{tabular}}
\end{table}

\begin{figure*}
\centering
 \includegraphics[width=1.0\linewidth]{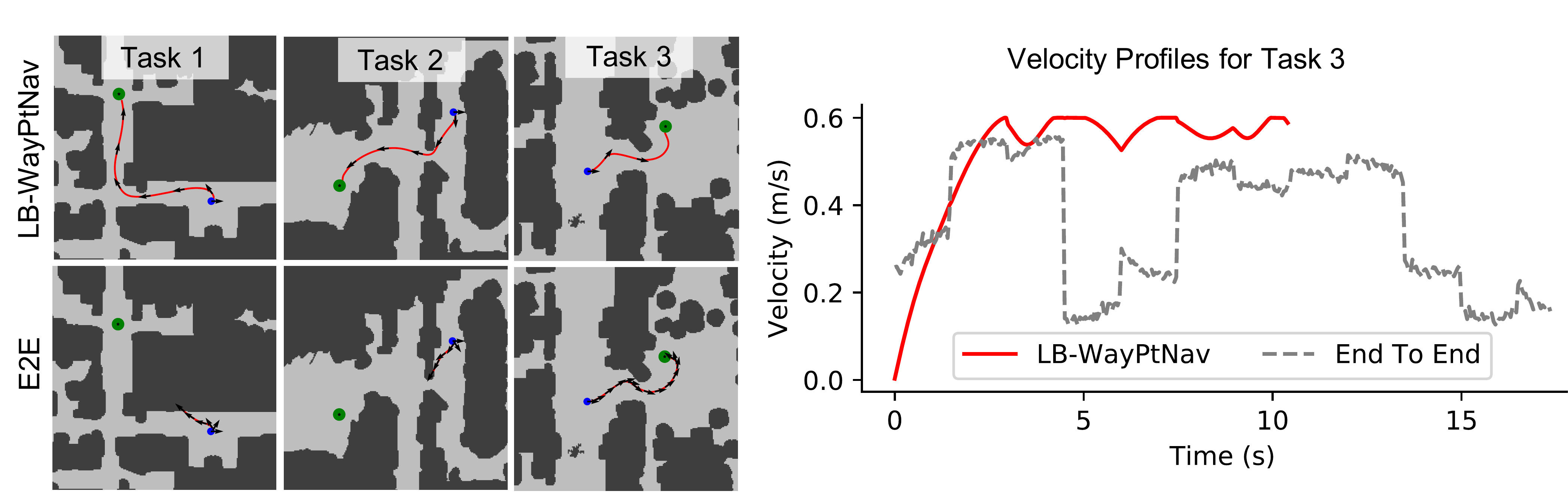}
\caption{\textbf{Trajectory Visualization:} We visualize the trajectories produced by the model-based planning approach (top row) and the end-to-end (E2E) learning approach (bottom row) for sample test tasks.
The E2E learning approach struggles to navigate around the tight corners or narrow hallways, whereas \metName{} is able to produce a smooth, collision-free trajectory to reach the target position.
Even though both approaches are able to reach the target position for task 3, \metName{} takes only 10s to reach the target whereas the E2E learning approach takes about 17s. 
Moreover, the control profile produced by the E2E learning approach is significantly more jerky than \metName{}, which is often concerning for real robots as they are power inefficient, can lead to significant errors in sensors and cause hardware damage.}
\label{fig:state_trajs_ours}
\end{figure*}

\subsection{Results}
\textbf{Comparison with the End-to-End learning approach.}
Table~\ref{table:success_bar} presents quantitative comparisons.
We note that \metName{} conveys the robot to the goal location 
more often (22\% higher success rate), much faster (40\% less 
time to reach the goal), and with less power 
consumption (50\% less acceleration).
Figure~\ref{fig:state_trajs_ours}(left) shows top-view visualization of 
trajectories executed by the two methods. Top-views
maps are being only used for visualization, both methods operate
purely based on first-person RGB image inputs.
As \metName{} uses a model-based planner to compute exact controls, it only
has to learn ``where to go'' next, as opposed to the E2E method that also
needs to learn ``how to go'' there. Consequently, \metName{}
is able to successfully navigate through narrow hallways, and 
make tight turns around obstacles and corners, while E2E method 
struggles.
This is further substantiated by the velocity control profiles 
in Figure \ref{fig:state_trajs_ours}(right). Even though the 
E2E method was trained to predict smooth control profiles (as generated 
by the expert policy), the control profiles at test time are still 
discontinuous and jerky. We also experimented with adding a smoothing
loss while training the E2E policy; though it helped reduce the 
average jerk, there was also a significant decline in the success rate.
This indicates that learning both an accurate and a smooth control profile
can be a hard learning problem. In contrast, as \metName{} uses model-based
control for computing control commands, it achieves average acceleration 
and jerk that is as low as that of an expert.\footnote{For a fair comparison, 
we report these metrics only over the test tasks that all approaches succeed at.}
This distinction has a significant implication for actual 
robots since the consumed power is directly proportional 
to the average acceleration. Hence, for the same battery capacity, \metName{}
will drive the robot twice as far as compared to E2E learning. 

\textbf{Comparison with Geometric Mapping and Planning Approach.}
We note that an online geometric mapping and planning approach, 
when used with ideal depth image observations, achieves near-expert 
performance. This is not surprising as perfect 
depth and egomotion satisfies the exact assumptions made
by such an approach.
Since \metName{} is a reactive planning framework, we also compare 
to a memory-less version that uses a map derived from 
\textit{only} the current depth image.
Even though the performance of memory-less planner is 
comparable to \metName{}, it still outperforms slightly 
due to the perfect depth estimation in simulation. 
However, since real-world depth sensors are neither 
perfect nor have an unlimited range, we see a noticeable 
drop in the performance of mapping-based planners in 
real-world experiments as discussed in real world experiments in
Section \ref{sec:exp}.

\begin{figure}
\centering
  \includegraphics[width=0.95\columnwidth]{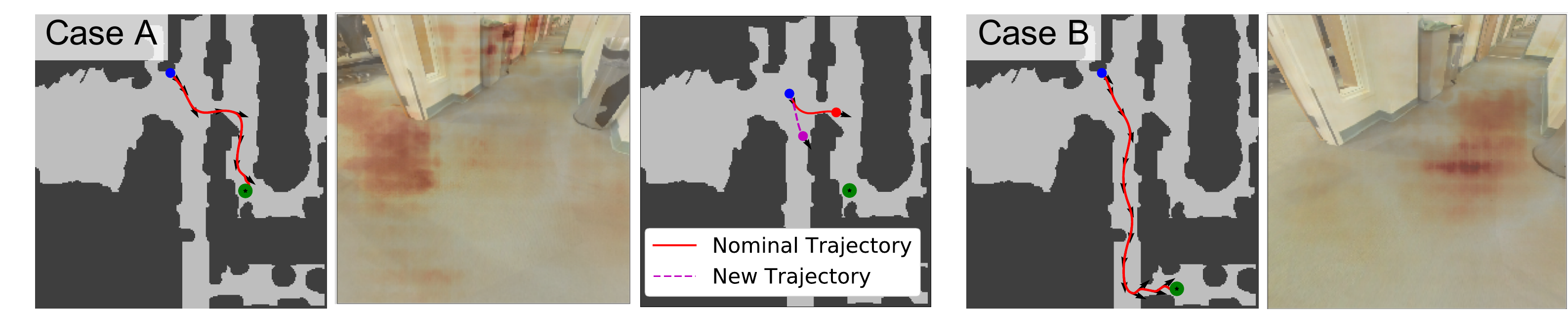}
    \caption{\metName{} is able to learn the appropriate navigation cues, such as entering the room through the doorway for a goal inside the room, continuing down the hallway for a farther goal. Such cues enable the robot to navigate efficiently in novel environments.}
    \label{fig:semantics}
\end{figure}

\textbf{Visualization of Learned Navigation Affordances.}
We conducted some analysis to understand what cues \metName{}
leverages in order to solve navigation tasks. 
Figure~\ref{fig:semantics} shows two related navigation tasks 
where the robot is initialized in the same state, but is tasked to either
go inside a close by room (Case A), or to a far away room that
is further down the hallway (Case B). \metName{} correctly
picks appropriate waypoints, and is able to reason that
a goal in a far away room is better achieved by going down the 
hallway as opposed to entering the room currently in front 
of the robot.

We also conduct an occlusion sensitivity analysis \cite{zeiler2014visualizing}, 
where we measure the change in the predicted waypoint as a rectangular
patch is zeroed out at different locations in the input image. 
We overlay the magnitude of this change in prediction on the input 
image in Red.
\metName{} focuses on the walls, doorways, hallways and obstacles such as trash-cans
as it predicts the next waypoint, and what the network attends to depends
on where the robot is trying to go. Furthermore, for Case A, we also show
the changed waypoint (in pink) as we zero out the wall pixels. This 
corresponds to a shorter path to the goal in the absence of the wall.
More such examples in the appendix in 
Section \ref{sec:appendix:semantics}. 

\textbf{Failure Modes.}
Even though \metName{} is able to perform navigation tasks 
in novel environments, it can only do local reasoning 
(there is no memory in the network) and gets stuck in some situations.
The most prominent failure modes are: a) when the robot 
is too close to an obstacle, and b) situations that require 
`backtracking' from an earlier planned path.

%% file: sections/experiments3.tex
\section{Hardware Experiments} \label{sec:exp}
We next test \metName{} on a TurtleBot 2 hardware testbed.\footnote{
More details about the hardware testbed in Sec. \ref{appendix_sec:hardware}. Experiment videos are in the supplementary material.}
We use the network trained in simulation, as described in 
Section \ref{sec:sims}, and deploy it directly on the 
TurtleBot without any additional training or finetuning.
We tested the robot in two different buildings, neither of 
which is in the training dataset (in fact, not even in the 
S3DIS dataset).\footnote{Representative images of our experiment environments are shown in Figure~\ref{fig:experiment_scenarios} 
in Section~\ref{sec:appendix:training_test_areas}.}
For state measurement, we use the on-board odometry sensors on the TurtleBot.
Test environments for the experiments 
are described in Table \ref{tab:experiment_setups}.

\input{sections/real-exp-desc4.tex}

\begin{table}
\caption{\textbf{Quantitative Comparisons for Hardware Experiments:}
We deploy \metName and baselines on a TurtleBot 
2 hardware testbed for four navigation tasks for 5 trials per task. We report the success rate (higher is better), average time to reach goal, jerk and acceleration along the robot trajectory (lower is better).}
\label{table:success_bar_real_world}
\centering
\resizebox{1.0\linewidth}{!}{
\begin{tabular}{lccccc}
\toprule
\textbf{Agent} & \textbf{Input} & \textbf{Success ($\%$)} & \textbf{Time taken ($s$)} & \textbf{Acceleration ($m/s^{2}$)} & \textbf{Jerk ($m/s^{3}$)} \\ \midrule
\metName{} (our) & RGB & 95 & 22.93 \textpm 2.38 &   0.09 \textpm 0.01 &   3.01 \textpm 0.38\\
End To End & RGB & 50 & 33.88 \textpm 3.01 &    0.19 \textpm 0.01 &    6.12 \textpm 0.18 \\
Mapping (memoryless)  & RGB-D & 0 &    N/A &    N/A &   N/A \\
Mapping & RGB-D + Spatial Memory & 40 &  22.13 \textpm 0.54 &    0.11 \textpm 0.01 &    3.44 \textpm 0.21 \\
\bottomrule
\end{tabular}}
\end{table}

We repeat each experiment for our method and the three 
baselines: E2E learning, mapping-based planner, and a memoryless 
mapping-based planner, for 5 trials each.
Results across all 20 trials are summarized in Table~\ref{table:success_bar_real_world}, where
we report success rate, time to reach the goal, acceleration and jerk.

\textbf{Comparison to E2E learning} are consistent with our 
conclusions from simulation experiments. \metName{} results in more 
reliable, faster, and smoother robot trajectories.

\textbf{Comparison to Geometric Mapping and Planning.} Geometric mapping 
and planning is implemented using the RTAB-Map
package~\cite{labbe2019rtab}. RTAB-Map uses RGB-D images as captured by an 
on-board RGB-D camera to output an occupancy map that is used
with the spline-based planner to output motor commands.
As our approach only uses the current image, we also report 
performance of a memory-less variant of this baseline where 
occupancy information is derived only from the current observation.
While \metName{} is able to solve $95\%$ of the trials, this
memory-less baseline completely fails. It tends to convey 
the robot too close to obstacles, and fails to recover.
In comparison, the map building scheme performs better, with a $40\%$ success rate.
This is still a lot lower than performance of our method ($95\%$),
and near perfect performance of this scheme in simulation.
We investigated the reason for this poor performance, and found 
that this is largely due to imperfections in depth measurements in the 
real world. For example, the depth sensor fails to pick-up shiny 
bike-frames and helmets, black bike-tires and monitor screens, and 
thin chair legs and power strips on the floor. These systematically
missing obstacles cause the robot to collide in experiment 1 and 2.
Map quality also substantially deteriorates in the presence of 
strong ambient light, such as when the sunlight comes in through the window
in experiment 4 (visualizations and videos in supplementary). 
These are known fundamental issues with depth sensors, that limit 
the performance of classical navigation stacks that crucially rely on them. 

\textbf{Performance of \metName:} 
In contrast, our proposed learning-based scheme that leverages
robot's prior experience with similar objects, operates
much better without need for extra instrumentation in the form of 
depth sensors, and without building explicit maps, for the 
short-horizon tasks that we considered. 
\metName is able to precisely control the robot through narrow hallways
with obstacles (as in experiment 1 and 2) while maintaining a smooth 
trajectory at all times. 
This is particularly striking, as the dynamics model used 
in simulation is only a crude approximation of the physics of a 
real robot (it does not include any mass and inertia 
effects, for example). The LQR feedback controller compensates for these approximations,
and enables the robot to closely track the desired trajectory 
(to an accuracy of $4cm$ in experiment 1).
\metName also successfully leverages 
navigation cues (in experiment 3 when it exits the room through a doorway), 
even when such a behavior was never hard-coded. Furthermore, thanks
to the aggressive data augmentation, \metName is able to perform well
even under extreme lighting conditions as in Experiment 4.

\begin{wrapfigure}{r}{0.25\textwidth}
 \vspace{-0.5cm}
  \begin{center}
    \includegraphics[width=\linewidth]{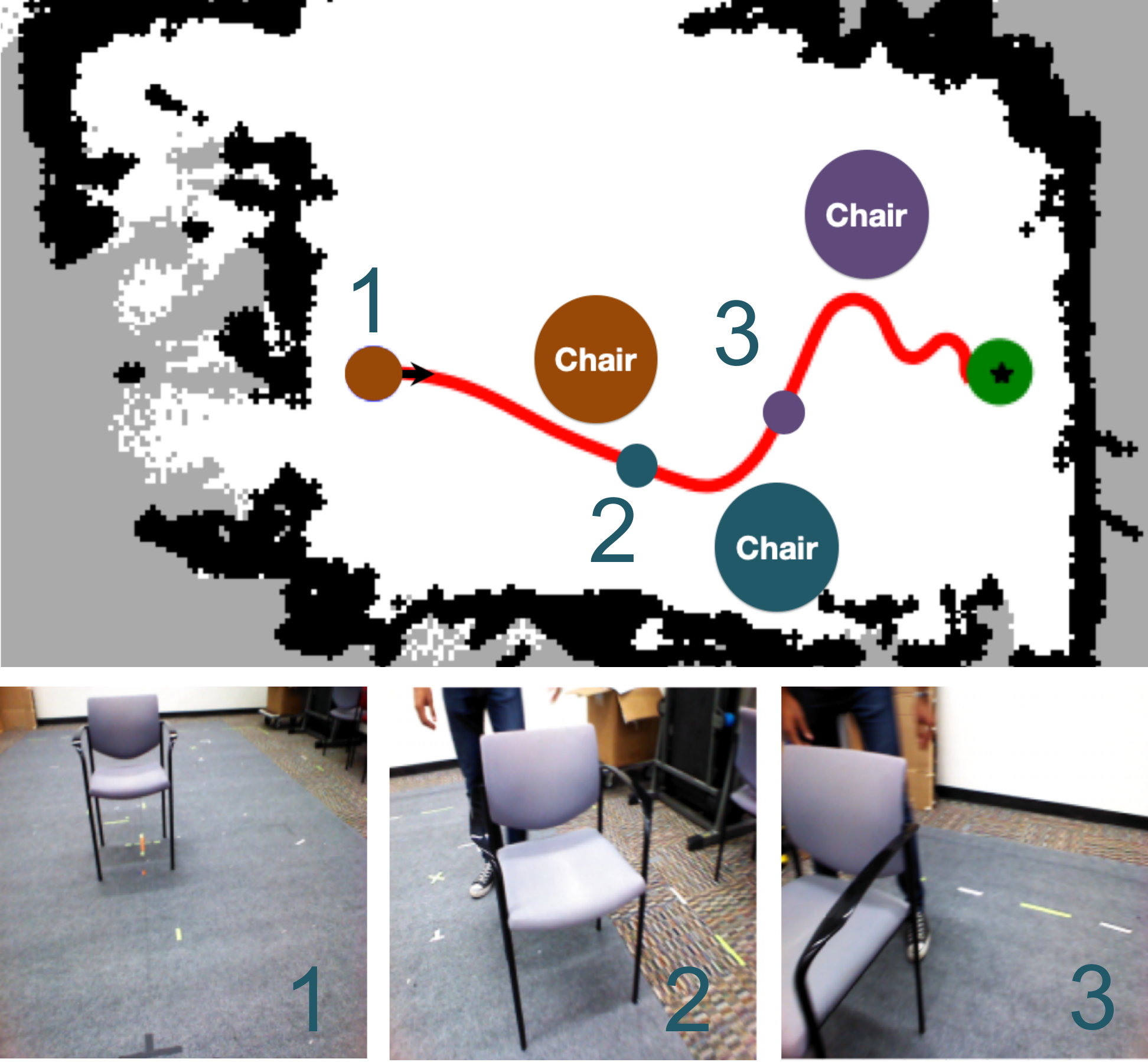}
  \end{center}
  \vspace{-0.3cm}
  \caption{\metName can adapt to dynamic environments.}
  \vspace{-1.8cm}
  \label{fig:dynamics}
\end{wrapfigure}
Furthermore, \metName is agile and reactive. It can adapt to 
changes in the environment. In an additional experiment 
(shown in Figure~\ref{fig:dynamics}), we change the environment as the 
robot executes its policy. The robot's goal is to go 
straight $6m$. It must go around the brown chair. 
As the policy is executed, we move the chair to repeatedly 
block the robot's path (we show the new chair locations 
in blue and purple, and mark the corresponding 
positions of the robot at which the chair was moved by 
same colors). We decrease the control horizon to $0.5s$ 
for this experiment to allow for a faster visual feedback.
The robot successfully reacts to the moving 
obstacle and reaches the target without colliding.

%% file: sections/real-exp-desc4.tex
\setlength{\tabcolsep}{3pt}
\begin{table}
\centering
\small
\caption{Experiment setups, with top-views (obtained offline 
only for visualization), and sample images.
Robot starts at the blue dot, and has to arrive at the green 
dot. Path taken by \metName is shown in red.}
\label{tab:experiment_setups}
\begin{tabular}{p{4.4cm}p{4.4cm}p{4.4cm}}
\toprule
\textbf{Experiment 1 and 2} & \textbf{Experiment 3} & \textbf{Experiment 4} \\
\midrule
\includegraphics[width=0.95\linewidth]{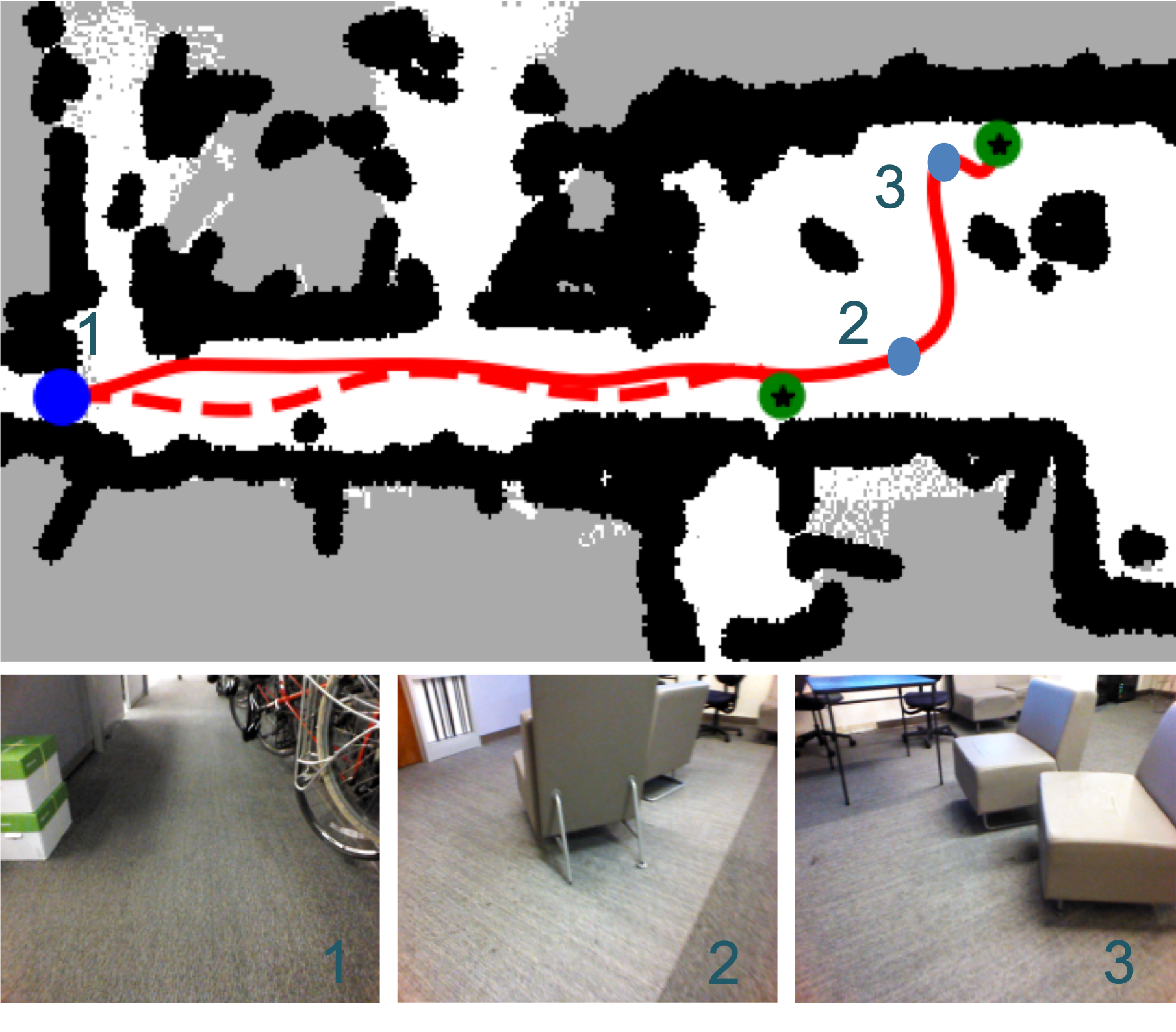} &
\includegraphics[width=0.95\linewidth]{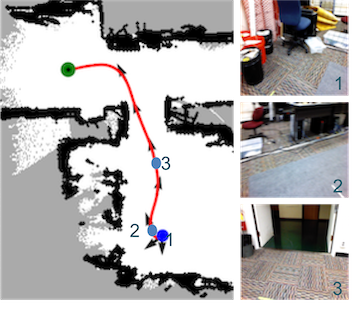} &
\includegraphics[width=0.95\linewidth, trim={0 0 0 2.4cm}, clip]{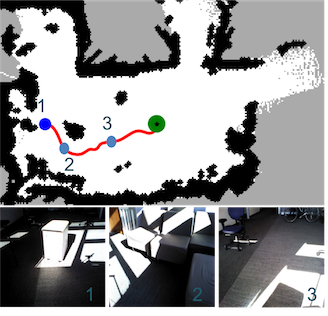} \\
\midrule
\textbf{Navigation through cluttered environments:} 
This tests if the robot can skillfully pass through clutter
in the real world: a narrow hallway with bikes on a bike-rack on one side,
and an open space with chairs and sofas. & 
\textbf{Leveraging navigation affordances:} 
This tests use of semantic cue for effective navigation.
Robot starts inside a room facing a wall.
Robot needs to realize it must exit the room through the doorway
in order to reach the target location. &
\textbf{Robustness to lighting conditions:} 
Experiment area is similar to that used for experiment 1, 
but experiment is performed during the day when sunlight comes 
from the windows. Robot needs to avoid 
objects to get to the goal. \\
\bottomrule
\end{tabular}
\end{table}

%% file: sections/conclusion.tex
\section{Discussion} \label{sec:conclusion}
We propose \metName, a navigation framework that combines learning and model-based control for goal-driven navigation in novel indoor environments. 
\metName is better and more reliable at reaching unseen goals compared to an End-to-End learning or a geometric mapping-based approach.
Use of a model-based feedback controller allows \metName to successfully generalize from simulation to physical robots.

Even though \metName{} is somewhat robust to the domain gap between simulation and real-world environments, when the appearances of objects are too different between the two, it fails to predict good waypoints. 
Thus, it might be desirable to finetune \metName{} using real-world data.
\metName{} also assumes perfect state estimation and employs a purely reactive policy.
These assumptions may not be optimal for long range tasks, wherein incorporating long-term spatial memory in the form of (geometric or learned such 
as in~\cite{gupta2017cognitive}) maps is often critical.
Furthermore, a more detailed study of dynamic environments would be an interesting future work.

%% file: sections/acknowledgment.tex
\section*{Acknowledgments} \label{sec:acknowledge}
This research is supported in part by the DARPA Assured Autonomy program under agreement number FA8750-18-C-0101, by NSF under the CPS Frontier project VeHICaL project (1545126), by NSF grants 1739816 and 1837132, by the UC-Philippine-California Advanced Research Institute under project IIID-2016-005, by SRC under the CONIX Center, and by Berkeley Deep Drive.

%% file: sections/appendix.tex
\section{Supplementary Material}

\subsection{Network Architecture and Training Details} \label{appendix_sec:arch}
\textbf{Implementation Details:}
We train \metName{} and E2E agents on 125K data points generated by our expert policy (Section \ref{appendix_sec:expert_data}). All our models are trained with a single GPU worker using TensorFlow~\cite{abadi2016tensorflow}. 
We use MSE loss on the waypoint prediction (respectively on the control command prediction) for training the CNN in our perception module (respectively for E2E learning).
We use Adam optimizer to optimize the loss function with a batch size of 64.
We train both networks for 35K iterations with a constant learning rate of $10^{-4}$ and use a weight decay of $10^{-6}$ to regularize the network.
We use ResNet-50~\cite{DBLP:journals/corr/HeZRS15}, pre-trained for ImageNet Classification, to encode the image input.
We remove the top layer, and use a downsampling convolution layer, followed by 5 fully connected layers with 128 neurons each to regress to the optimal waypoint (or control commands for E2E learning).
The image features obtained at the last convolution layer are concatenated with the egocentric target position and the current linear and angular speed before passing them to the fully connected layers (see Figure \ref{fig:framework}).
During training, the ResNet layers are finetuned along with the fully connected layers to learn the features that are relevant to the navigation tasks.
We use standard techniques used to avoid overfitting including dropout following each fully connected layer except the last (with dropout probability $20\%$) \cite{srivastava2014dropout}, and data augmentation such as randomly distorting brightness, contrast, adding blur, perspective distortion at training time \cite{simard2003best}.
Adding these distortions significantly improves the generalization capability of our framework to unseen environments. More details about the kinds of distortions and their effect on the performance can be found in Section \ref{sec:appendix:image_distortions}.

\subsection{Expert supervision} \label{appendix_sec:expert_data}
To generate supervision for training the perception network, we use a Model Predictive Control (MPC) scheme to find a sequence of dynamically feasible waypoints and the corresponding spline trajectories that navigate the robot from the starting state to the goal state.
This can be done during the training phase because a map of the environment is available during the training time.
However, no such privileged information is used during the test time. 

To generate the expert trajectory, we define a cost function that trades-off the distance from the target position and the nearest obstacle, and optimize for the waypoint such that the resultant spline trajectory to the waypoint minimizes this cost function.
More specifically, given the map of the environment, we compute the signed distance function to the obstacles, $d^{obs}(x, y)$, at any given position $(x, y)$. 
Given the goal position, $\pos^*$, of the vehicle, we compute the minimum collision-free distance to the goal, $d^{goal}(x, y)$ (also known as the FMM distance to the goal). 
The overall cost function to the MPC problem is then given by:
\begin{align}
\label{eqn:expert_cost}
    J(\mathbf{z}, \mathbf{u}) = & \sum_{i=0}^{T} J_i(z_i, u_i), \\
    J_i(z_i, u_i) := &\left(\max\{0, \lambda_1 - d^{obs}(x_i, y_i)\}\right)^3 + \lambda_2 \left(d^{goal}(x, y)\right)^2,
\end{align}
where $\mathbf{z} := (z_0, z_1, \ldots, z_T)$ is the state trajectory, $\mathbf{u}$ is the corresponding control trajectory, $T$ is the maximum time horizon, and $\lambda_2$ trades-off the distance from the goal position and the obstacles. 
We only penalize for the obstacle cost when the corresponding robot trajectory is within a distance of $\lambda_1$ to an obstacle. 
Moreover, the obstacle cost is penalized more heavily compared to the goal distance (a cubic penalty vs a quadratic penalty). 
This is done to ensure that the vehicle trajectory does not go too close to the obstacles. 
We empirically found that it is significantly harder to learn to accurately predict the waypoints when the vehicle trajectory goes too close to the obstacles, as the prediction error margin for the neutral network is much smaller in such a case, leading to a much higher collision rate.
Given the cost function in \eqref{eqn:expert_cost}, the overall MPC problem is given as
\begin{align}
& \min_{\mathbf{z}, \mathbf{u}} J(\mathbf{z}, \mathbf{u}) \label{eqn:expert_mpc_problem} \\
\text{subject to~~} & x_{i+1} = x_i + \Delta T v_i \cos\phi_i,\quad y_{i+1} = y_i +  v_i\sin\phi_i,\quad \phi_{i+1} = \phi_{i} + \Delta T \omega_i\,, \label{eqn:expert_mpc_problem_help1}\\
& v_i \in [0, \bar{v}],\quad \omega_i \in [-\bar{\omega}, \bar{\omega}]\,, \label{eqn:expert_mpc_problem_help2}\\
& z_0 = (0, 0, 0), \quad u_0 = (0, 0), 
\end{align}
where $\Delta T$ is the discretization step for the dynamics in \eqref{eqn:NumSimpleDyn}, and the initial state and speed are assumed to be zero without loss of generality.

Starting from $i=0$, we solve the MPC problem in \eqref{eqn:expert_mpc_problem} in a receding horizon fashion. 
In particular, at any timestep $i=t$, we find a waypoint such that the corresponding spline trajectory respects the dynamics and the control constraints in \eqref{eqn:expert_mpc_problem_help1} and \eqref{eqn:expert_mpc_problem_help2} (and hence is a dynamically feasible trajectory), and minimizes the cost in \eqref{eqn:expert_mpc_problem} over a time horizon of $\horizon_1$. Thus, we solve the following optimization problem:
\begin{align}
& \min_{\waypt_t} \sum_{i=t}^{t+H_1} J_i(z_i, u_i) \label{eqn:expert_mpc_problem2}\\
\text{subject to~~} & \{z, u\}_{t:t+H_1} = FitSpline(\waypt_t, u_t),\\
& z_t, u_t \text{~~- Given} \label{eqn:expert_mpc_problem2_help2}
\end{align}
where $\waypt_t := (\hat{x}_{\time}, \hat{y}_{\time}, \hat{\theta}_{\time})$ is the waypoint, and $\{z, u\}_{t:t+H_1}$ are the corresponding state and control spline trajectories that satisfy the dynamics and the control constraints in \eqref{eqn:expert_mpc_problem_help1} and \eqref{eqn:expert_mpc_problem_help2}, and respect the boundary conditions on the trajectories imposed by $z_t$, $u_t$ and $\waypt_t$. 
Such spline trajectories can be computed for a variety of aerial and ground vehicles (see \cite{koo1999differential, mellinger2011minimum} for more details on the aerial vehicles and \cite{WALAMBE2016601} for the ground vehicles).
Thus, the feasible waypoints must be reachable from the current state and speed of the vehicle while respecting the vehicle's control bounds.
For example, a pure rotation waypoint (i.e., $\waypt_t = (0, 0, \hat{\theta}))$ is not feasible for the system if it has a non-zero linear speed at time $t$, and thus will not be considered as a candidate solution of \eqref{eqn:expert_mpc_problem2}.
These dynamics considerations are often ignored in the learning-based methods in literature which work with a set of macro-actions.
This often results in an undesirable, jerky, and stop-and-go behavior on a real robot.
In this work, we use third-order polynomial splines, whose coefficients are computed using the values of $z_t$, $u_t$ and $\waypt_t$ (see \cite{WALAMBE2016601} for more details).
Use of third-order splines make sure that the velocity and acceleration profiles of the vehicle are smooth, and respect the control bounds.

Let the optimal waypoint corresponding to the optimization problem in \eqref{eqn:expert_mpc_problem2} be $\waypt_t^*$, and the corresponding optimal trajectories be $\{z^*, u^*\}_{t:t+H_1}$. 
The image obtained at state $z^*_t$, $I_t$, the relative goal position $p^*_t$, the speed of the robot $u^*_t$, and the optimal waypoint $\waypt_t^*$ thus constitutes one data point for the training. 
For the End-to-End learning, we use $u^*_{t:t+H}$ instead of $\waypt_t^*$ for training the network.

Given the low dimension of the waypoint (three in our case), we use a sampling-based approach to compute the optimal waypoint in \eqref{eqn:expert_mpc_problem2}.
We sample the waypoints within the ground-projected field-of-view of the vehicle (ground projected assuming no obstacles).
Note, even though we use a sampling-based approach to obtain the optimal waypoint, other gradient-based optimization schemes can also be used, especially when the state space of the vehicle is high-dimensional.
We next apply the control sequence $u^*$ for the time horizon $[t, t+\horizon]$ to obtain the state $z^*_{t+\horizon}$, and repeat the entire procedure in \eqref{eqn:expert_mpc_problem2} starting from time $t+\horizon$. 
We continue this process until the robot reaches the goal position.
In our work, we use $\lambda_1 = 0.3m, \Delta T = 0.05s, H_1=6s$ and $H=1.5s$.  

The procedure outlined in this section allows us to compute large training datasets using optimal control without requiring any explicit human labeling. 
Moreover, the generated waypoints are guaranteed to satisfy the dynamics and control constraints.
Finally, the cost function in \eqref{eqn:expert_cost} allows us to explicitly ensure that the robot trajectory to the waypoint is collision-free, which is crucial in cluttered indoor environments. 
It is also worthwhile to note that this procedure can be applied to a variety of ground vehicles and aerial vehicles that are differentially flat.
For differentially flat systems, the path planning can be done with respect to a much lower-dimensional state space (or waypoint) using spline trajectories \cite{koo1999differential, WALAMBE2016601, mellinger2011minimum}, which makes the path planning tractable in real-time.

\begin{figure}[h!]
    \centering
    \begin{minipage}[c]{0.3\textwidth}
     \includegraphics[width=\columnwidth]{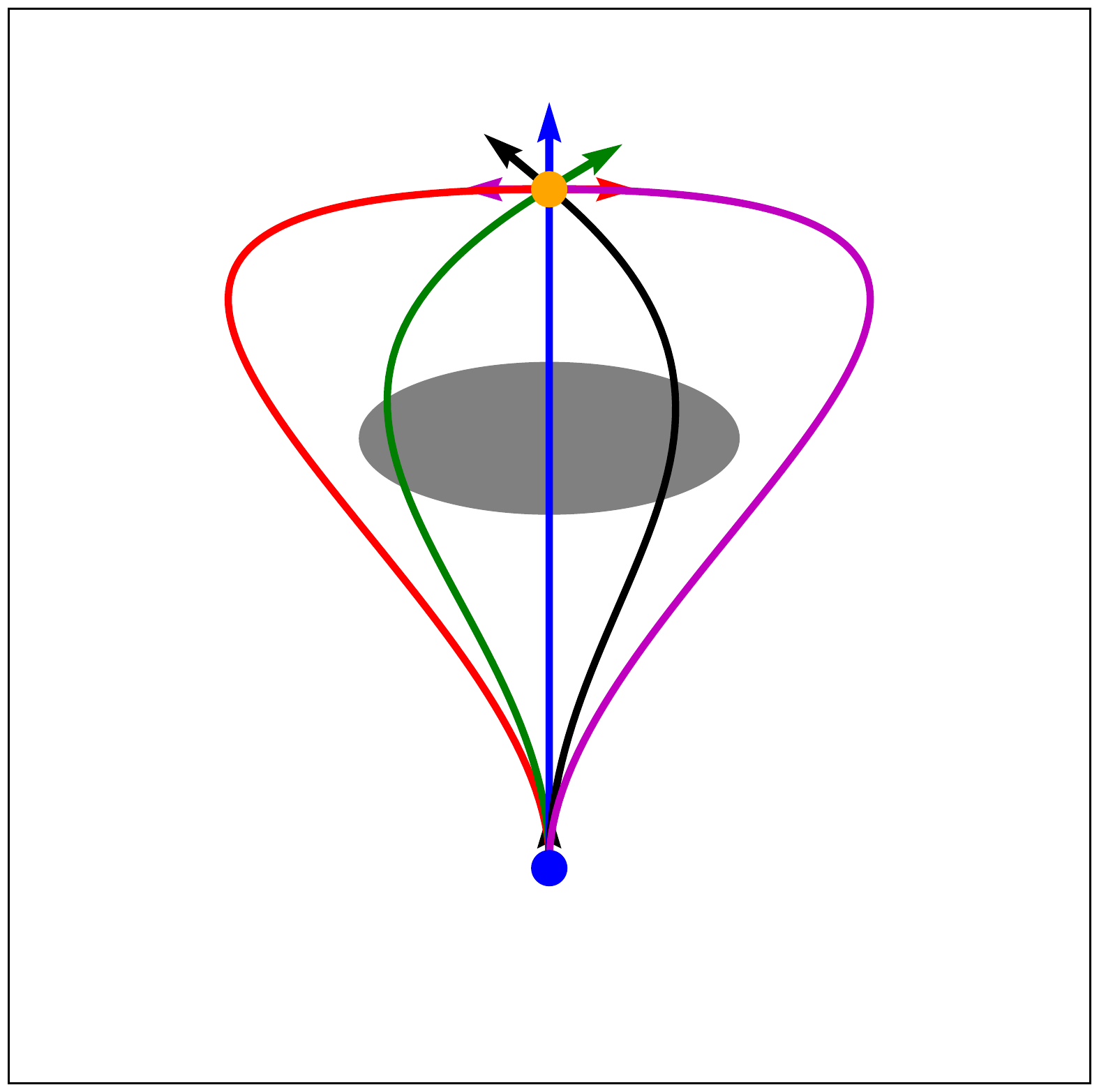}
     \end{minipage}\hfill
     \begin{minipage}[c]{0.6\textwidth}
        \caption{We visualize the spline trajectories produced by our planner for different waypoint angles ($\hat{\theta}$), starting from the same initial state, initial speed and to the same waypoint position.
        The gray region denotes an obstacle.
        Due to the dynamics constraints, $\hat{\theta}$ significantly affects the shape of the vehicle trajectory, and hence needs to be chosen appropriately to obtain a collision-free trajectory. In particular, the line of sight trajectory to the waypoint (the Blue trajectory) leads to a collision in this case, whereas if $\hat{\theta}$ is chosen appropriately, a smooth, agile trajectory that goes around the obstacle can be obtained.}
        \label{fig:angle_effect}
    \end{minipage}
\end{figure}

It is also important to note that intuitively the waypoint in our framework summarizes the local obstacle information in the environment, and hence, its representation choice is very crucial.
For example, the choice of $\hat{\theta}$ in $\waypt$ affects the shape of the trajectory the vehicle takes to the waypoint (see Figure \ref{fig:angle_effect}).
Typically, $\hat{\theta}$ is chosen as the line of sight angle between the robot's current position and the desired position $(\hat{x}, \hat{y})$ \cite{kaufmann2018deep, muller2018driving}.
However, $\hat{\theta}$ is an extra degree-of-freedom that can be chosen appropriately to generate rich, agile and collision-free trajectories in cluttered environments.

\newpage
\subsection{Image distortions and domain randomization} \label{sec:appendix:image_distortions}
During training, we apply a variety of random distortions to images, such as adding blur, removing some pixels, adding some superpixels, changing image saturation, sharpness, contrast and brightness. 
We also apply perspective distortions to images, such as varying the field-of-view and tilt (pitch) of the camera.
For adding these distortions, we use the \textit{imgaug} python library.

\begin{figure}[h!]
\centering
\begin{subfigure}[b]{0.24\columnwidth}
\centering
  \includegraphics[width=\columnwidth]{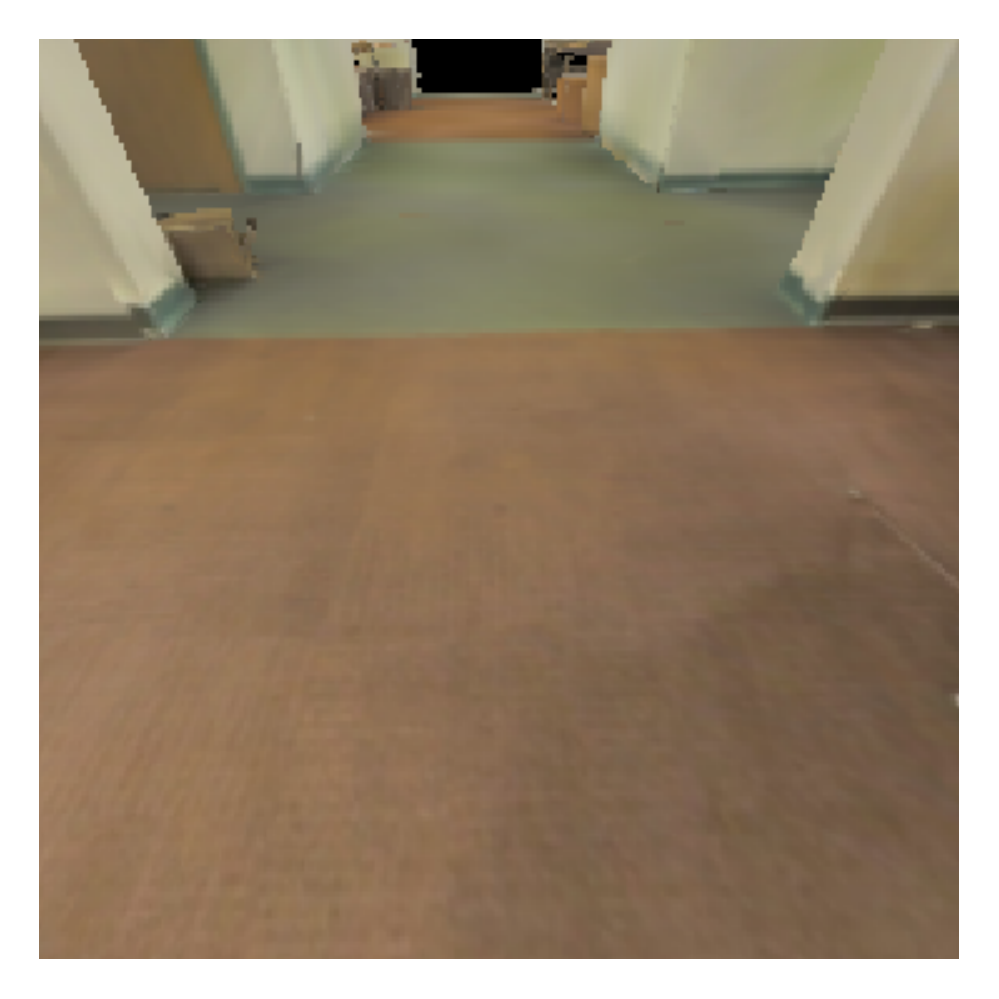}
  \subcaption{Undistorted image}
\end{subfigure}%
\hfill
\begin{subfigure}[b]{0.24\columnwidth}
\centering
  \includegraphics[width=\columnwidth]{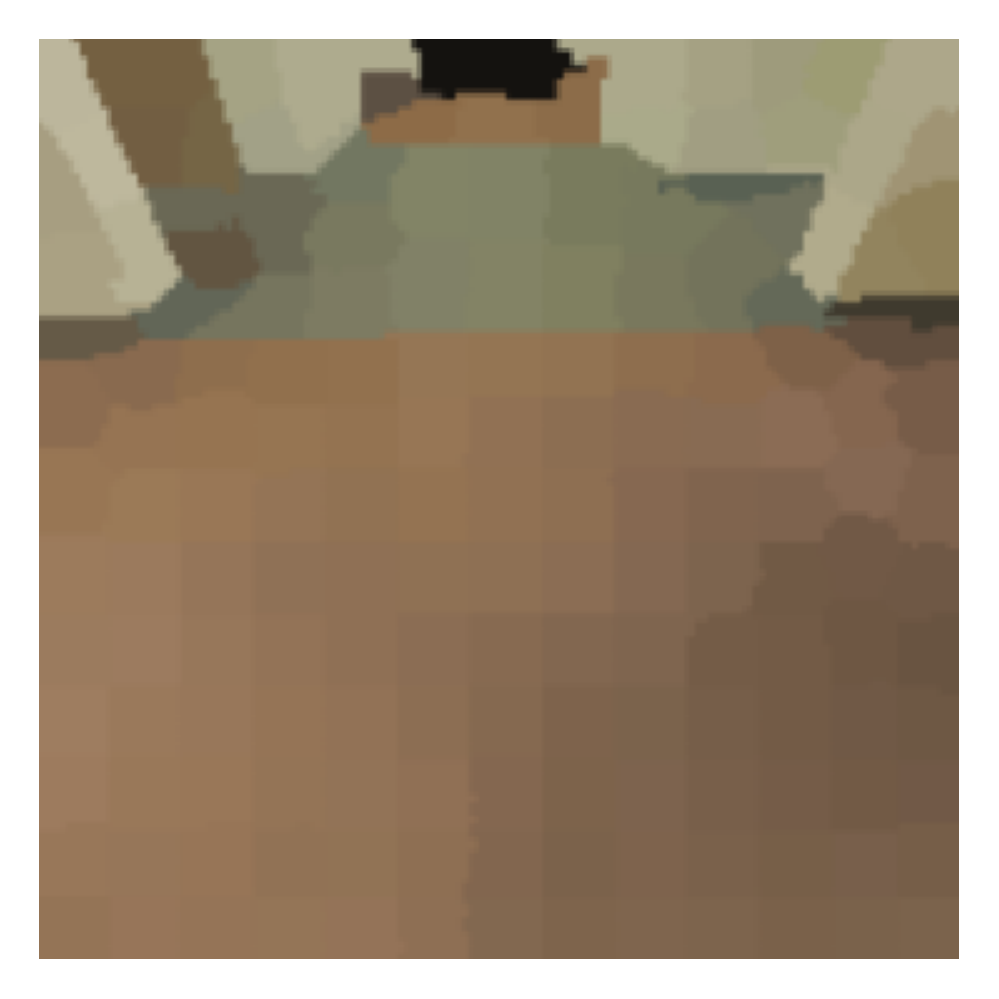}
  \subcaption{Adding superpixels}
\end{subfigure}
\hfill
\begin{subfigure}[b]{0.24\columnwidth}
\centering
  \includegraphics[width=\columnwidth]{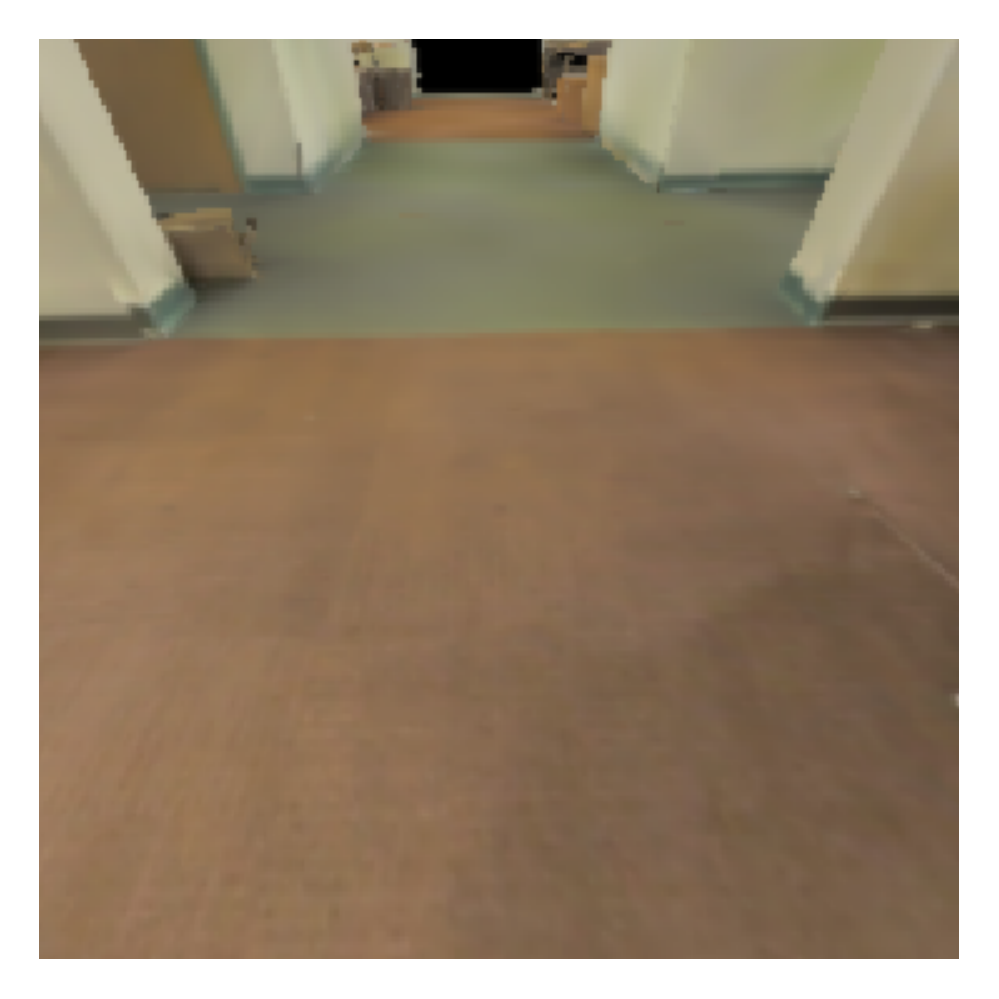}
  \subcaption{Adding Gaussian blur}
\end{subfigure}
\hfill
\begin{subfigure}[b]{0.24\columnwidth}
\centering
  \includegraphics[width=\columnwidth]{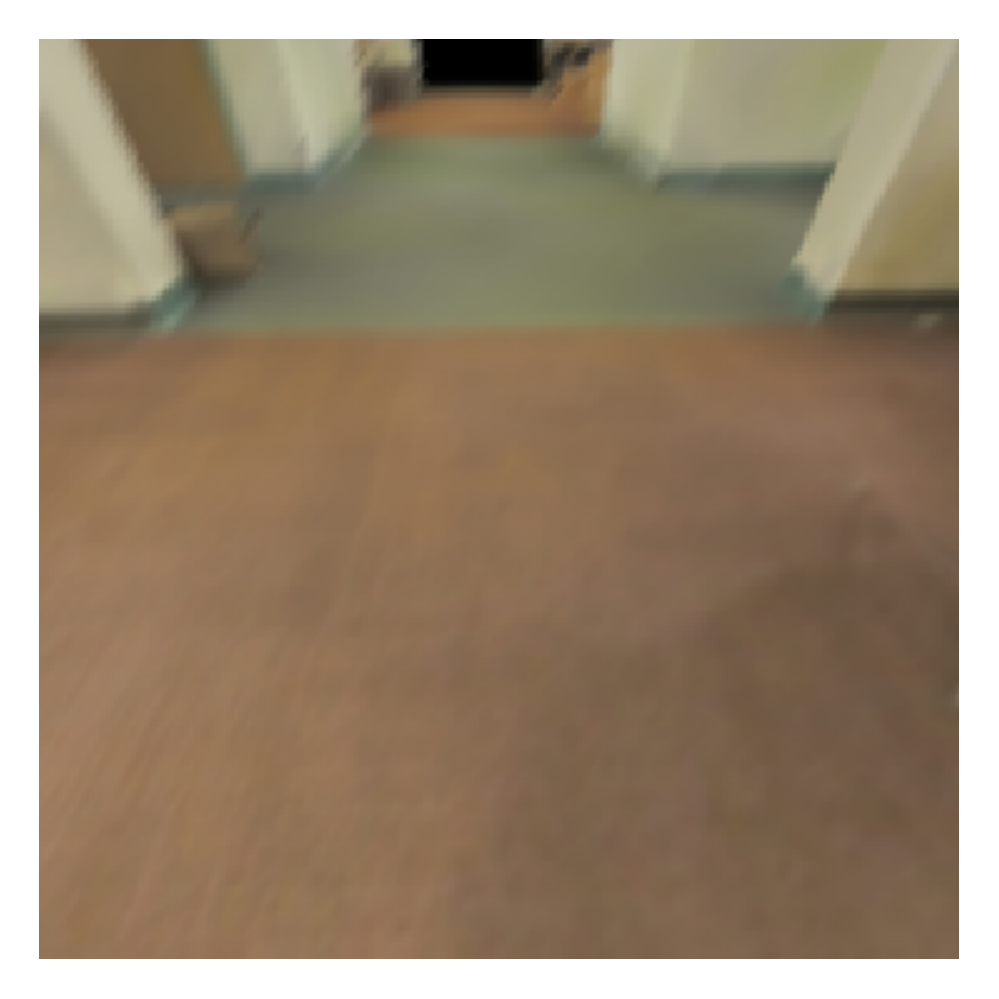}
  \subcaption{Adding motion blur}
\end{subfigure}

\begin{subfigure}[b]{0.24\columnwidth}
\centering
  \includegraphics[width=\columnwidth]{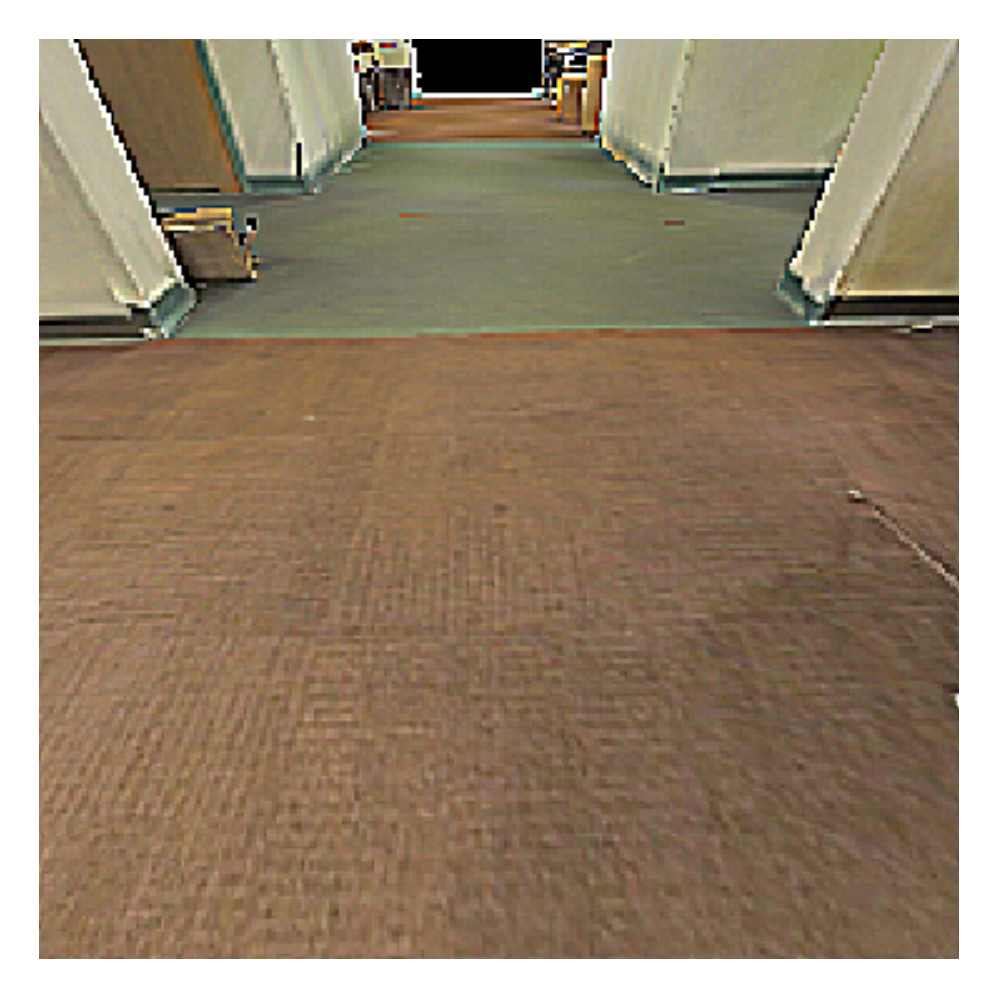}
  \subcaption{Image sharpening}
\end{subfigure}
\hfill
\begin{subfigure}[b]{0.24\columnwidth}
\centering
  \includegraphics[width=\columnwidth]{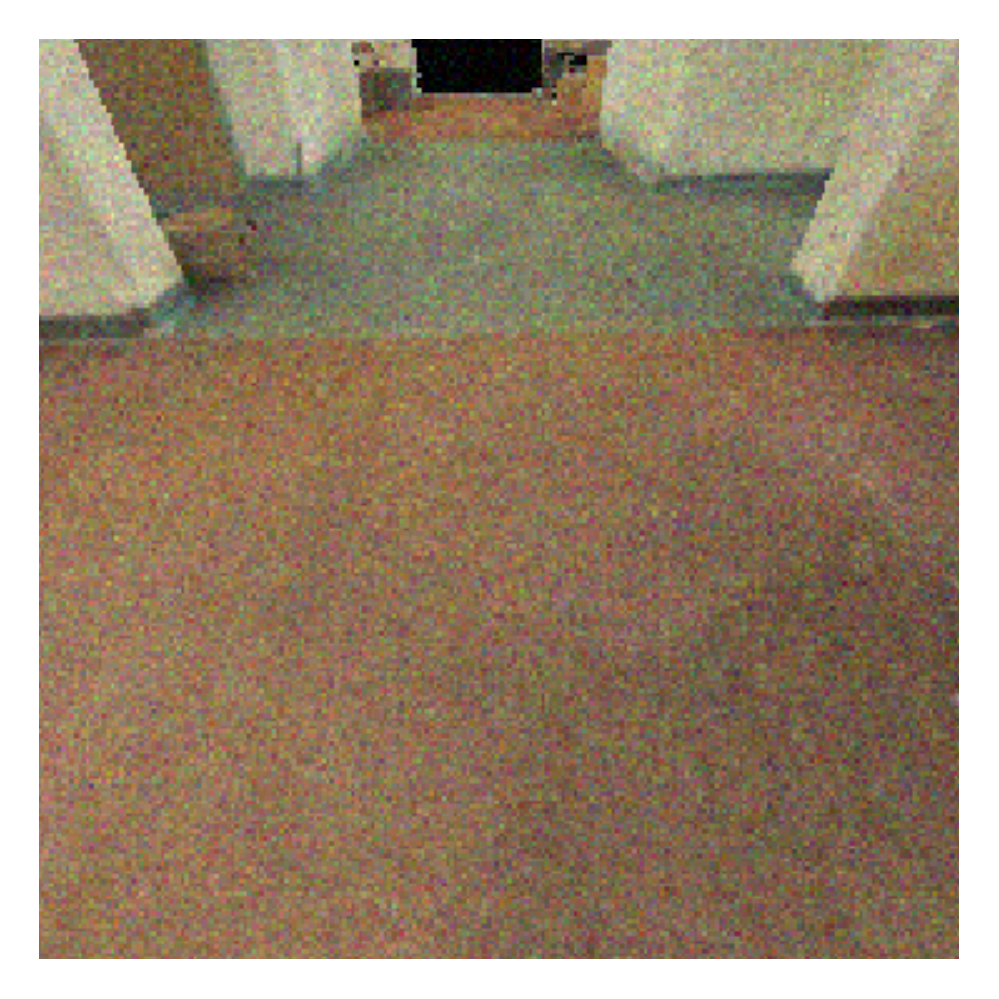}
  \subcaption{Adding Gaussian noise}
\end{subfigure}%
\hfill
\begin{subfigure}[b]{0.24\columnwidth}
\centering
  \includegraphics[width=\columnwidth]{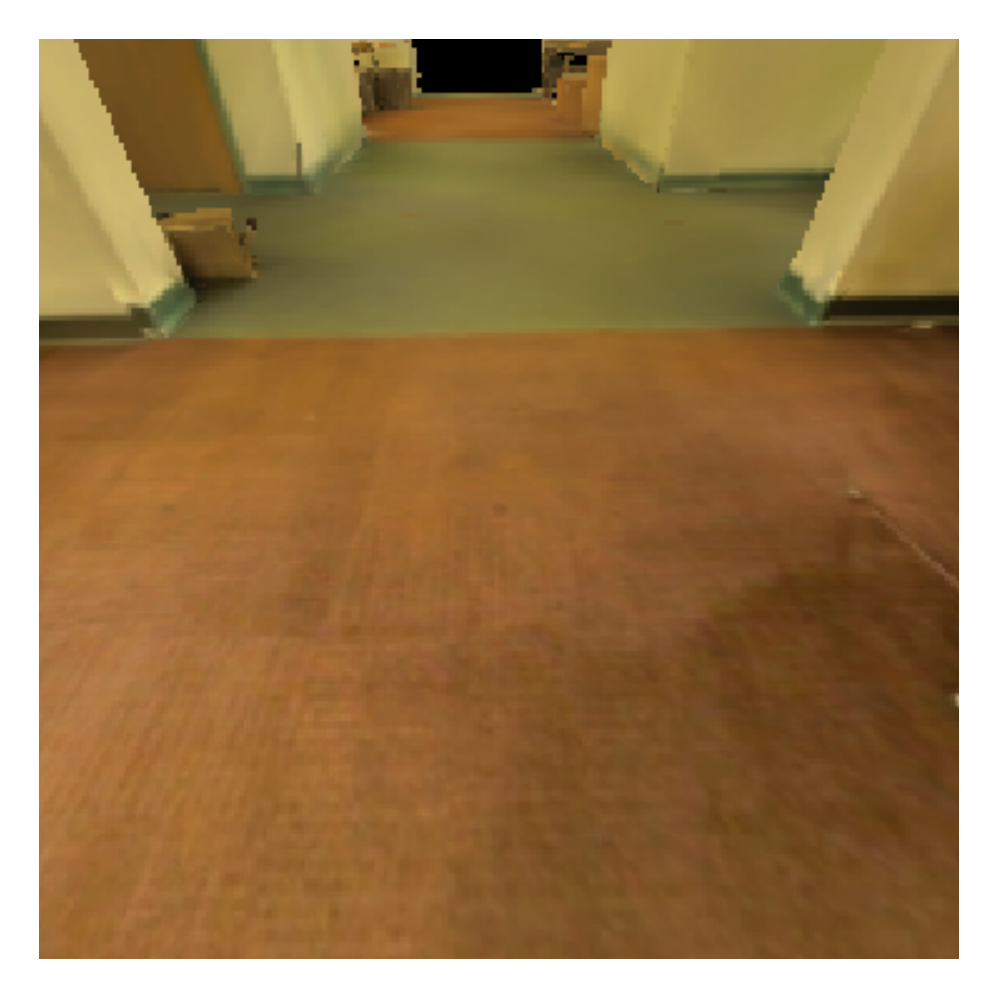}
  \subcaption{Changing brightness}
\end{subfigure}
\hfill
\begin{subfigure}[b]{0.24\columnwidth}
\centering
  \includegraphics[width=\columnwidth]{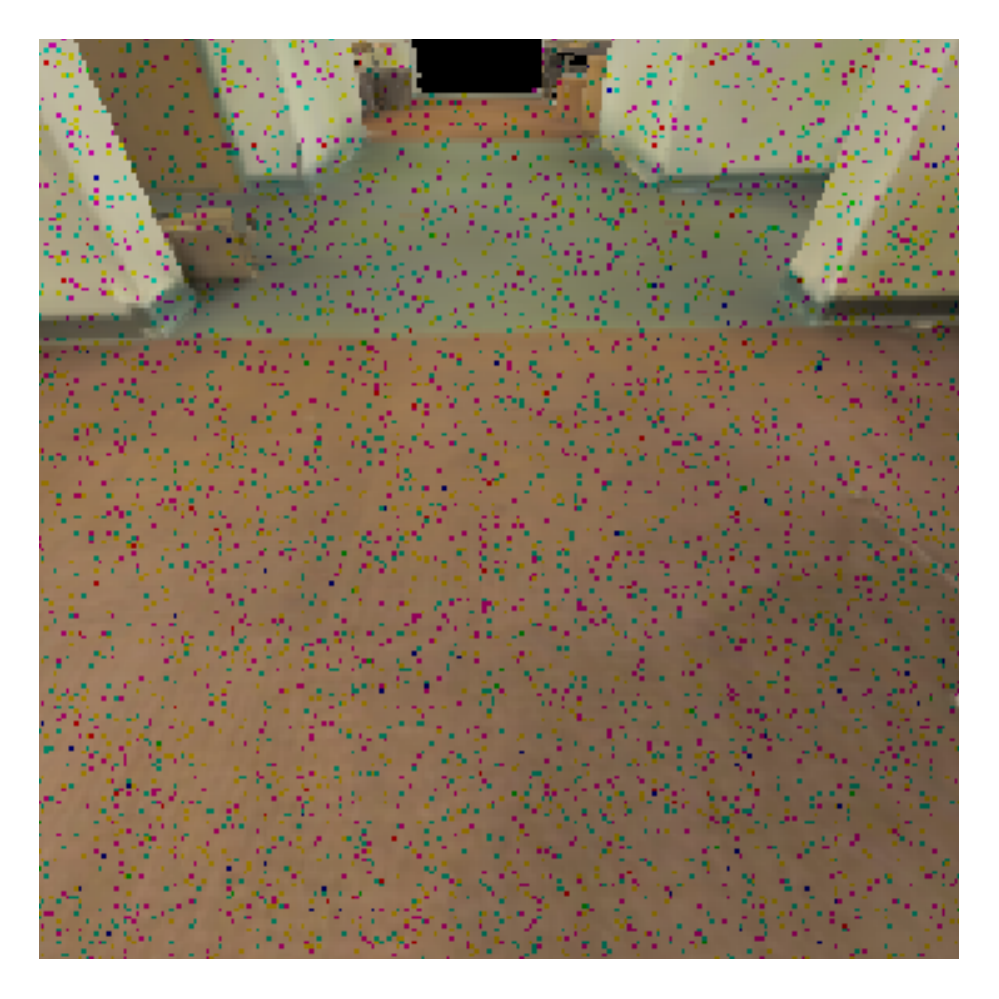}
  \subcaption{Dropping pixels}
\end{subfigure}

\begin{subfigure}[b]{0.24\columnwidth}
\centering
  \includegraphics[width=\columnwidth]{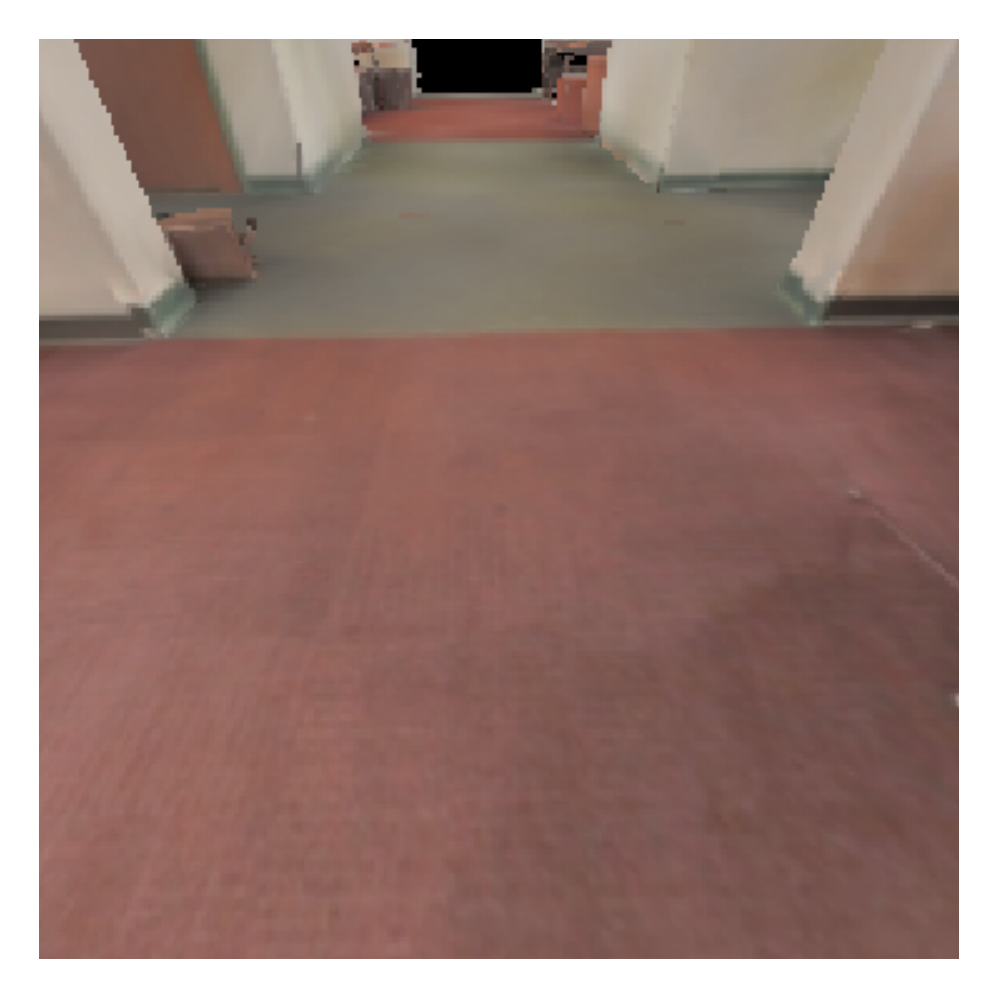}
  \subcaption{Changing saturation}
\end{subfigure}
\hfill
\begin{subfigure}[b]{0.24\columnwidth}
\centering
  \includegraphics[width=\columnwidth]{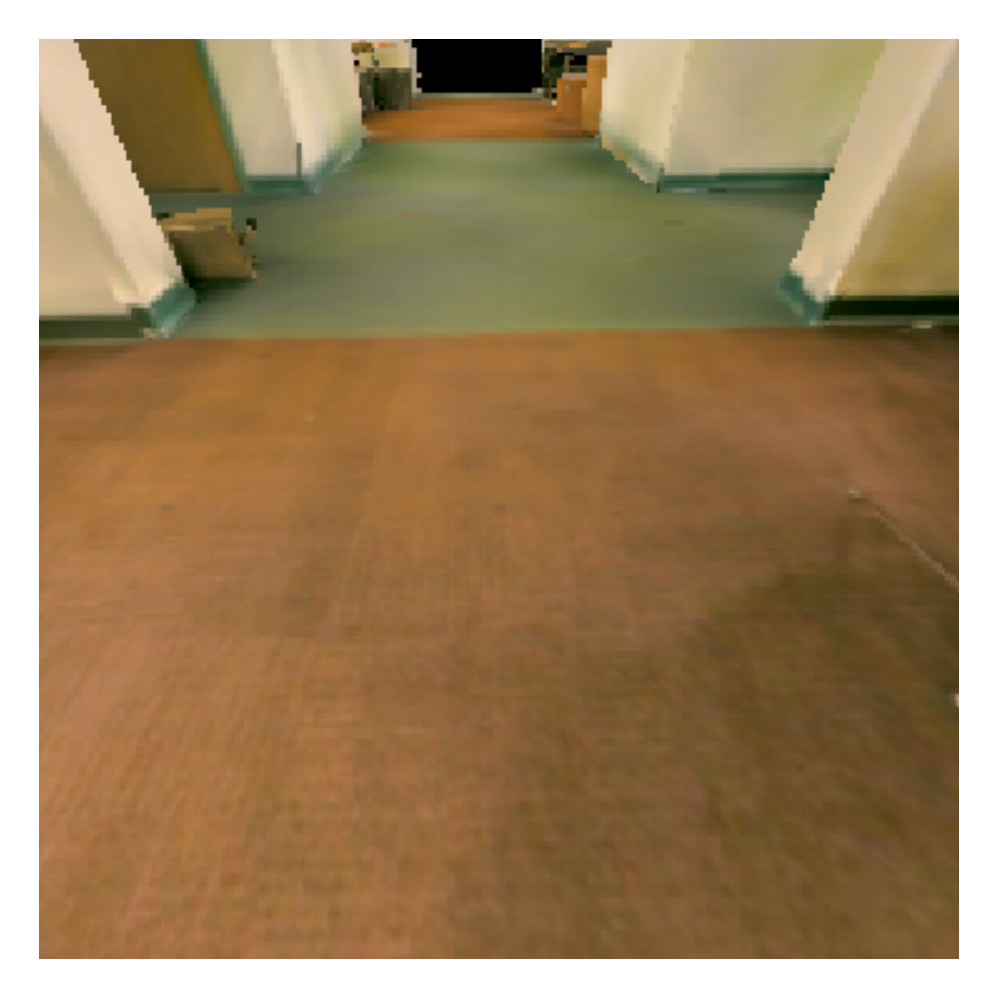}
  \subcaption{Changing contrast}
\end{subfigure}
\hfill
\begin{subfigure}[b]{0.24\columnwidth}
\centering
  \includegraphics[width=\columnwidth]{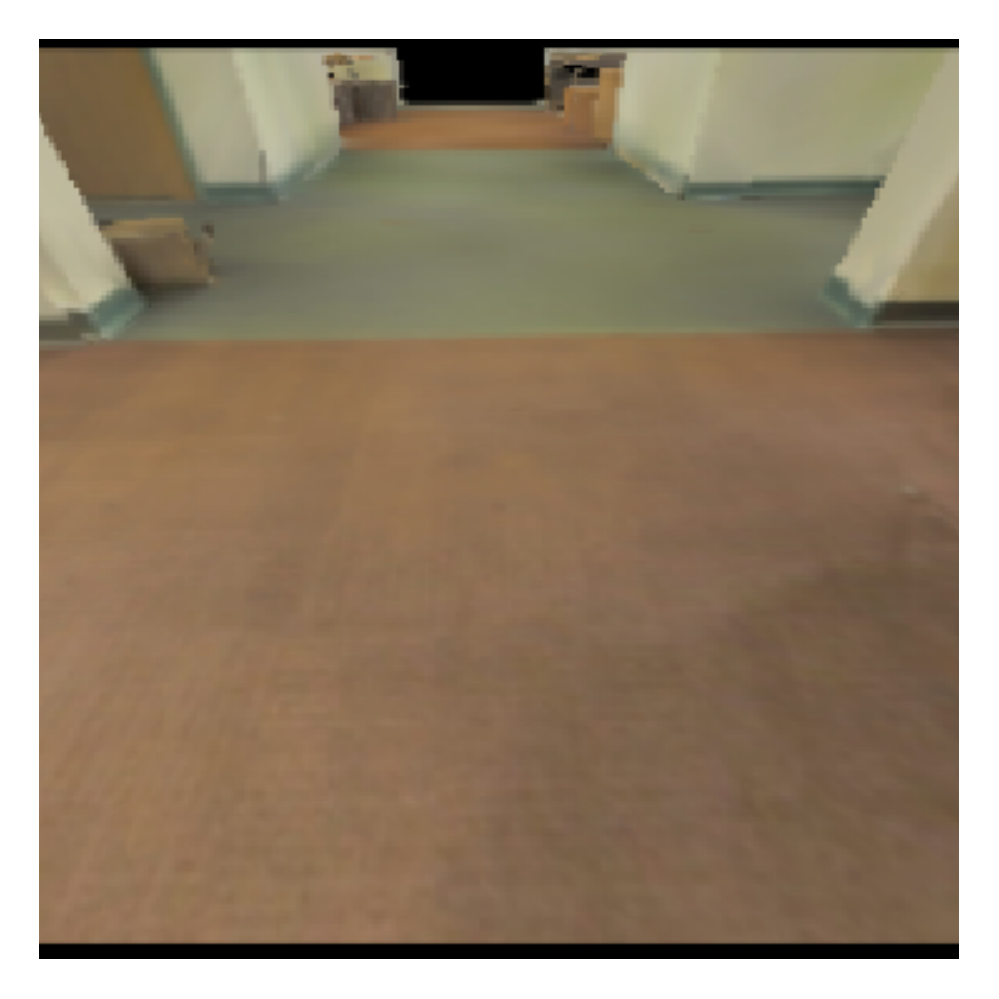}
  \subcaption{Changing field-of-view}
\end{subfigure}%
\hfill
\begin{subfigure}[b]{0.24\columnwidth}
\centering
  \includegraphics[width=\columnwidth]{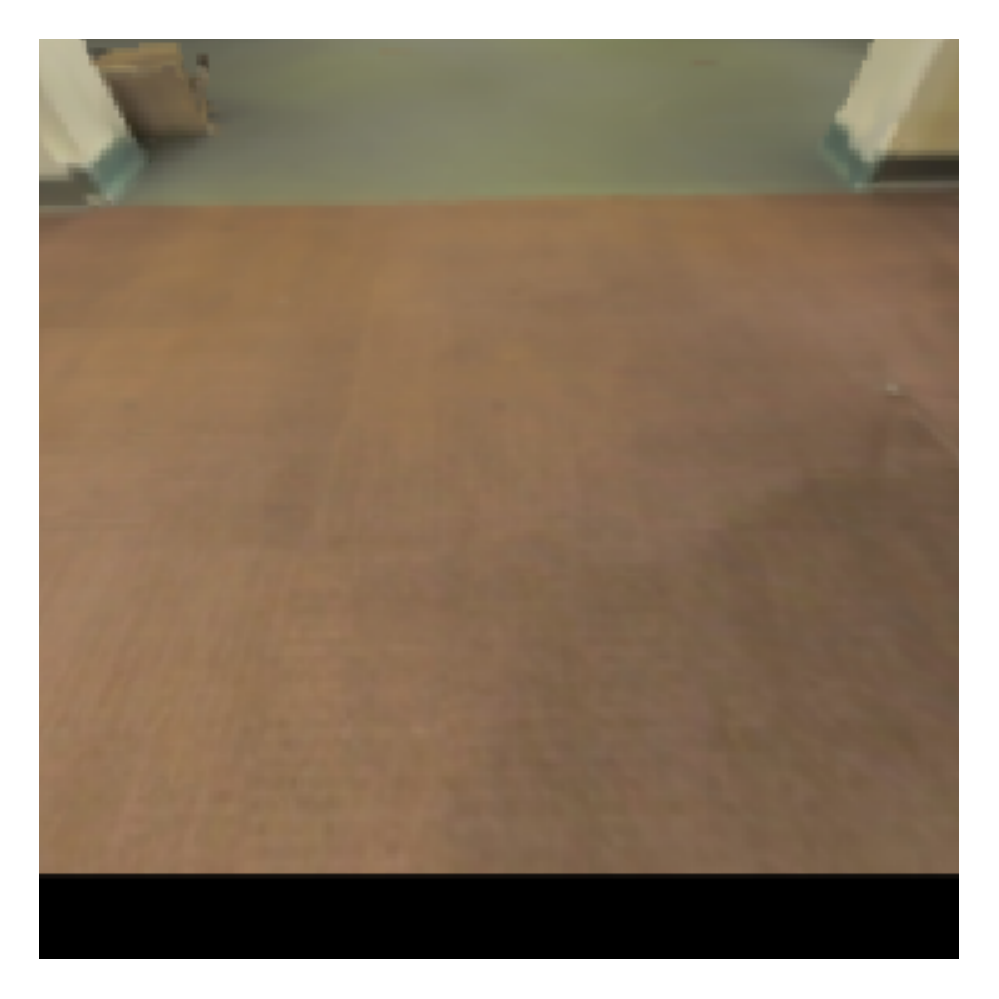}
  \subcaption{Changing camera tilt}
\end{subfigure}
\caption{Examples of several image distortions that have been randomly applied during the training phase. The actual undistorted image is shown in (a). Adding these distortions significantly improves the generalization capability of our framework to unseen environments.}
\label{fig:distortions}
\end{figure}

Some examples of sample distortions are shown in Figure \ref{fig:distortions}.
Adding these distortions during training significantly improves the generalization of the trained network to unseen environments.
For example, for \metName{}, adding distortions increases the success rate from 47.94\% to 80.65\%.
Adding perspective distortions particularly improves the generalization to the real-world systems, for which the camera tilt will inevitably change as the robot is moving through the environment. 

\newpage
\subsection{Training and test areas} \label{sec:appendix:training_test_areas}
\begin{figure}[h!]
\centering
\begin{subfigure}[b]{\columnwidth}
\centering
\begin{subfigure}[b]{0.24\columnwidth}
\centering
  \includegraphics[width=\columnwidth]{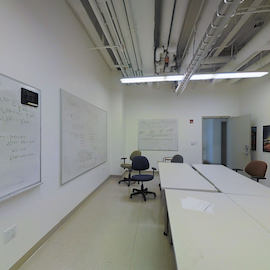}
\end{subfigure}%
\hfill
\begin{subfigure}[b]{0.24\columnwidth}
\centering
  \includegraphics[width=\columnwidth]{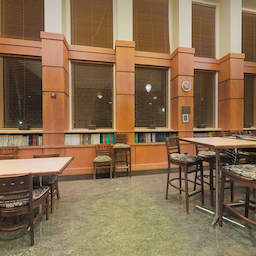}
\end{subfigure}
\hfill
\begin{subfigure}[b]{0.24\columnwidth}
\centering
  \includegraphics[width=\columnwidth]{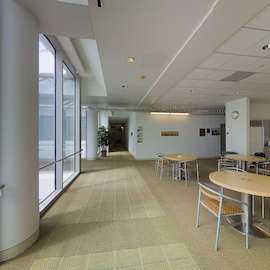}
\end{subfigure}
\hfill
\begin{subfigure}[b]{0.24\columnwidth}
\centering
  \includegraphics[width=\columnwidth]{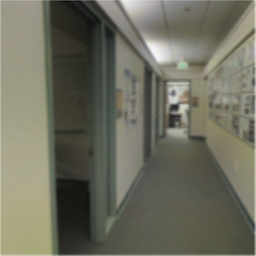}
\end{subfigure}
\caption{Training}
\end{subfigure}

\begin{subfigure}[b]{\columnwidth}
\centering
    \begin{subfigure}[b]{0.24\columnwidth}
    \centering
      \includegraphics[width=\columnwidth]{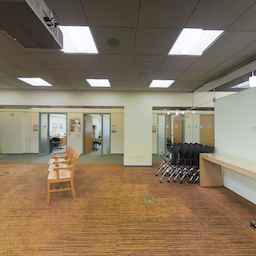}
    \end{subfigure}%
    \hfill
    \begin{subfigure}[b]{0.24\columnwidth}
    \centering
      \includegraphics[width=\columnwidth]{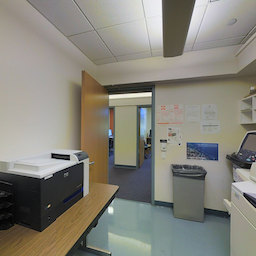}
    \end{subfigure}
    \hfill
    \begin{subfigure}[b]{0.24\columnwidth}
    \centering
      \includegraphics[width=\columnwidth]{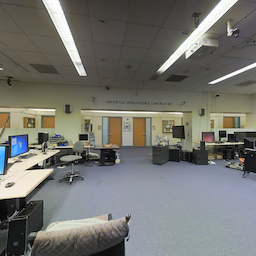}
    \end{subfigure}
    \hfill
    \begin{subfigure}[b]{0.24\columnwidth}
    \centering
      \includegraphics[width=\columnwidth]{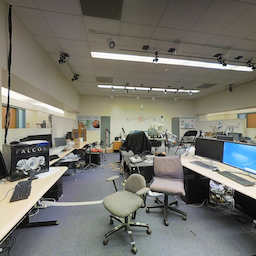}
    \end{subfigure}
    \caption{Test}
\end{subfigure}
\caption{Some representative images of the buildings from which the training and the test data was collected. Even though the test environments are also office buildings, their layouts and appearances are different than the training buildings. However, our framework is still able to generalize to the domain shift.}
\label{fig:training_test_areas}
\end{figure}

We illustrate some representative scenes from the test and the training buildings from the S3DIS dataset in Figure \ref{fig:training_test_areas}.
Note that the navigation tasks were not limited to these scenes and are spread across the entire building. 
Even though the layouts and appearances of the test buildings are different than the training buildings, our framework is able to adapt to the domain shift.

\begin{figure}[h!]
\centering
\begin{subfigure}[b]{0.24\columnwidth}
\centering
  \includegraphics[width=\columnwidth]{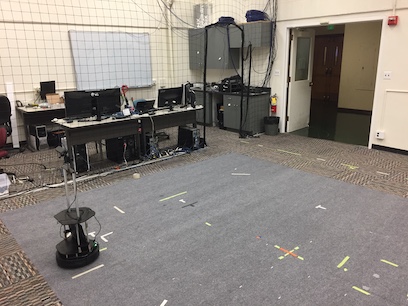}
\end{subfigure}%
\hfill
\begin{subfigure}[b]{0.24\columnwidth}
\centering
  \includegraphics[width=\columnwidth]{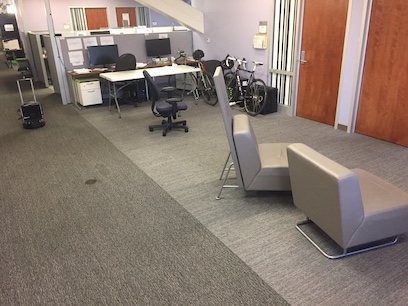}
\end{subfigure}%
\hfill
\begin{subfigure}[b]{0.24\columnwidth}
\centering
  \includegraphics[width=\columnwidth]{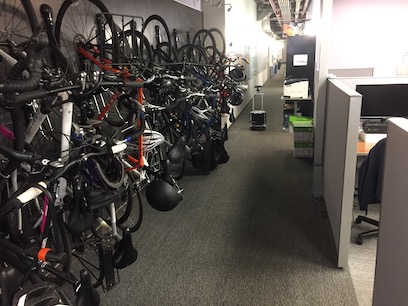}
\end{subfigure}%
\hfill
\begin{subfigure}[b]{0.24\columnwidth}
\centering
  \includegraphics[width=\columnwidth]{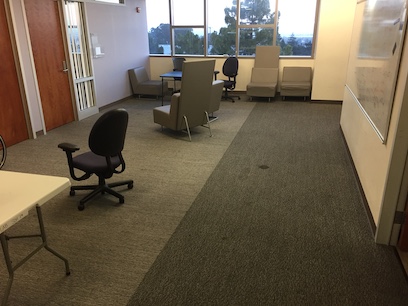}
\end{subfigure}%
\caption{Some representative images of the buildings in which the experiments were conducted. None of these buildings were used for training/testing purposes in simulation.}
\label{fig:experiment_scenarios}
\end{figure}

In our experiments, we test the robot in two different buildings, none of which is in the training dataset (in fact, not even in the S3DIS dataset).
We show some representative images of our experiment environments in Figure \ref{fig:experiment_scenarios}.

\newpage
\subsection{Learning Navigation Affordances} \label{sec:appendix:semantics}
\begin{figure}[h!]
\centering
\begin{subfigure}[b]{0.18\columnwidth}
\centering
  \includegraphics[width=\columnwidth]{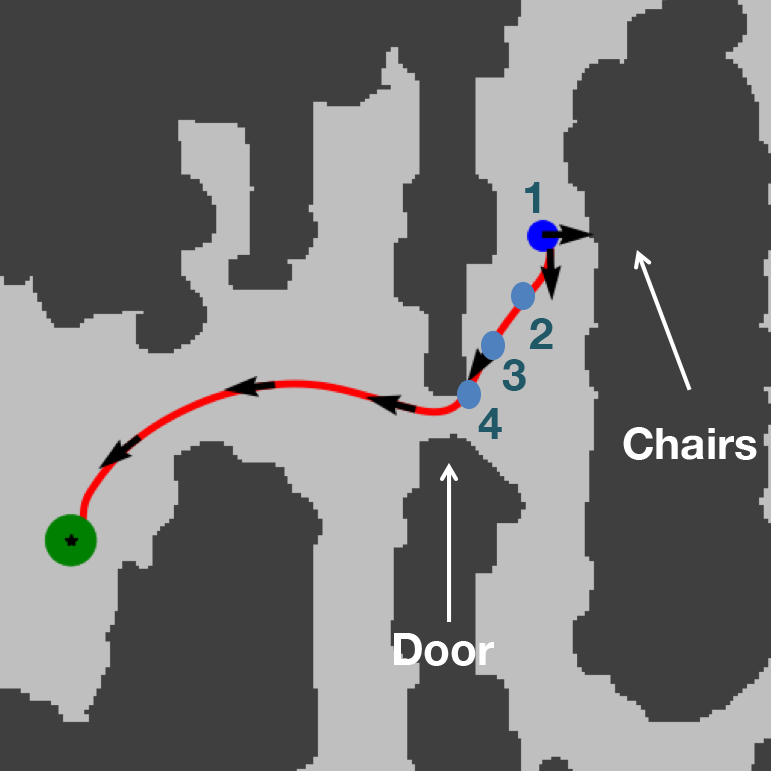}
  \subcaption{Top view (for vis. only)}
  \label{fig:semantics_ex1_1}
\end{subfigure} \hspace{1pt}
\begin{subfigure}[b]{0.18\columnwidth}
\centering
  \includegraphics[width=\columnwidth]{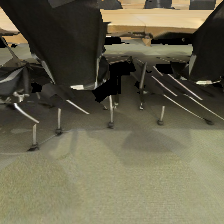}
  \subcaption{Image at Position 1}
  \label{fig:semantics_ex1_2}
\end{subfigure} \hspace{1pt}
\begin{subfigure}[b]{0.18\columnwidth}
\centering
  \includegraphics[width=\columnwidth]{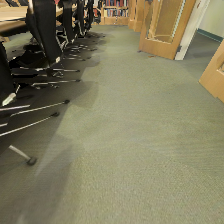}
  \subcaption{Image at Position 2}
  \label{fig:semantics_ex1_3}
\end{subfigure} \hspace{1pt}
\begin{subfigure}[b]{0.18\columnwidth}
\centering
  \includegraphics[width=\columnwidth]{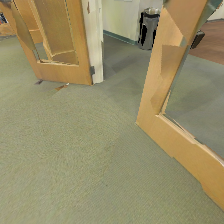}
  \subcaption{Image at Position 3}
  \label{fig:semantics_ex1_4}
\end{subfigure} \hspace{1pt}
\begin{subfigure}[b]{0.18\columnwidth}
\centering
  \includegraphics[width=\columnwidth]{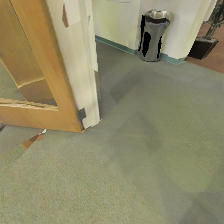}
  \subcaption{Image at Position 4}
  \label{fig:semantics_ex1_5}
\end{subfigure}%
    \caption{We visualize the trajectory as well as the observed RGB images (at the marked points) for \metName for a sample test task: the robot needs to go from the current room into another room. 
    Our method is able to learn the navigation cue of exiting the room through the doorway to get to the goal location. Such cues enable the robot to do a more efficient and guided exploration in novel environments.
    }
    \label{fig:semantics_ex1}
\end{figure}

In this section, we visualize some additional test cases and demonstrate how the proposed approach is indeed able to learn navigation cues for an efficient navigation in novel environments. 
In the first test case, the robot starts inside a conference room and is required to go to a target position that is in a different room.
The problem setup is shown in Figure \ref{fig:semantics_ex1_1}. 
The blue dot represents the robot's start position and the black arrow represents the robot's heading.
The center of the green area represents the target position. 

We mark some intermediate points on the robot trajectory in Figure \ref{fig:semantics_ex1_1} where it is required to predict a waypoint, and visualize the corresponding RGB images that it observed in Figures \ref{fig:semantics_ex1_2}-\ref{fig:semantics_ex1_5}.
The robot is initially facing some chairs (Fig. \ref{fig:semantics_ex1_2}), and based on the observed image, it predicts a waypoint to rotate in place to explore the environment.
Upon rotation, the robot sees part of the door (Fig. \ref{fig:semantics_ex1_3}) and is able to learn that to go to the target position it needs to exit the current room through this door and predicts a waypoint to go towards the door (Fig. \ref{fig:semantics_ex1_4}).
Next, it produces a waypoint to go out of the door (Fig. \ref{fig:semantics_ex1_5}), and eventually reaches the target position.
We also show the full trajectory (the red plot) of the robot from the start position to the target position as well as mark the obstacles in context in Figure \ref{fig:semantics_ex1_1}.

\begin{figure}[h!]
    \centering
\begin{subfigure}[c]{0.25\textwidth}
\centering
  \includegraphics[height=1.0\columnwidth]{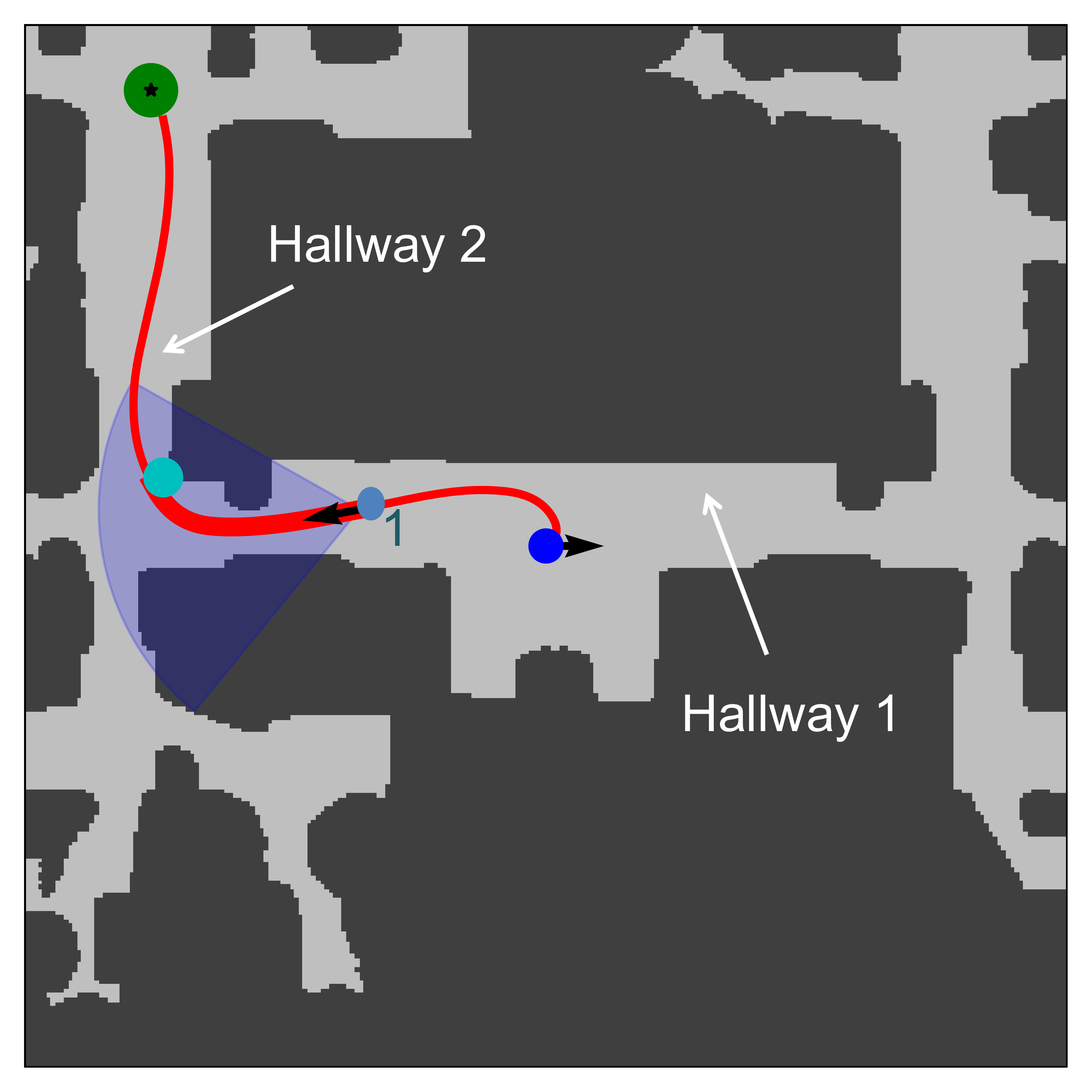}
  \subcaption{Top view (for vis. only)}
  \label{fig:semantics_ex2_1}
\end{subfigure} \hspace{1pt}
\begin{subfigure}[c]{0.25\textwidth}
\centering
  \includegraphics[height=1.0\columnwidth]{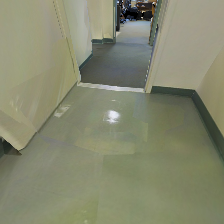}
  \subcaption{Image at Position 1}
  \label{fig:semantics_ex2_2}
\end{subfigure} \hspace{1pt}
     \begin{minipage}[c]{0.46\textwidth}
         \caption{We visualize the trajectory as well as the observed RGB images (at the marked point) for our method (\metName) for a sample test task. 
    In this task, the robot needs to go from one hallway (Hallway 1) into another (Hallway 2).
    Even though the entire Hallway 2 is not explicitly visible, the robot is able to reason that a hallway typically leads into another hallway, and uses this learned prior to efficiently reach its target position.}
    \label{fig:semantics_ex2}
    \end{minipage}
\end{figure}

As a second example, we consider the test case in Fig. \ref{fig:semantics_ex2_1} where the robot needs to go from Hallway 1 to Hallway 2.
We show an intermediate point in the robot trajectory (Marked 1 in Fig. \ref{fig:semantics_ex2_1}) where the robot needs to turn right into Hallway 2 to reach the target.
We also show the corresponding (stylized) field-of-view (projected on the ground) of the robot (the blue cone in Fig. \ref{fig:semantics_ex2_1}).
The corresponding RGB image that the robot observed is shown in Figure \ref{fig:semantics_ex2_2}.
Based on the observed image, the robot produces the cyan waypoint and the bold red desired trajectory.
Even though the robot is not able to see its next desired position (it is blocked by the door in the field-of-view), it can see part of Hallway 2 and is able to reason that a hallway typically leads to another hallway.
The robot uses this learned semantic prior to produce a waypoint leading into Hallway 2 and eventually reaches the target.

\newpage
\subsection{Hardware testbed} \label{appendix_sec:hardware}
\begin{wrapfigure}{r}{0.3\textwidth}
\vspace{-1.0cm}
  \begin{center}
    \includegraphics[width=\linewidth]{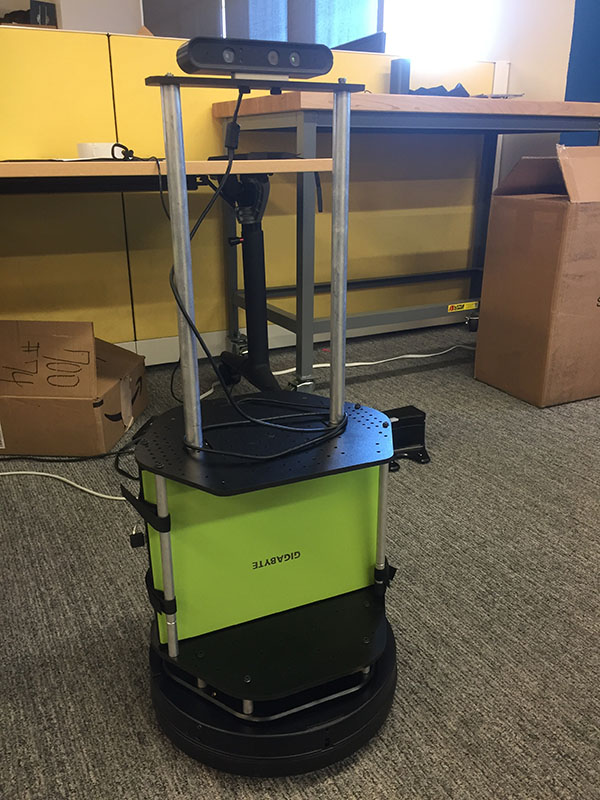}
  \end{center}
  \caption{Our Turtlebot 2 hardware platform uses a Yujin Kobuki base, Gigabyte Aero Laptop, and Orbbec Astra camera.}
  \vspace{-1cm}
  \label{fig:turtlebot}
\end{wrapfigure}
We use a TurtleBot 2 platform with a Yujin Kobuki robot base to serve as our autonomous vehicle during hardware experiments. 
The TurtleBot 2 is a low-cost, open source differential drive robot, which we equip with an Orbbec Astra RGB-D camera.
However, for \metName and E2E learning experiments, we only used the RGB image.
For geometric mapping and planning-based schemes, we additionally use the depth image.
A snapshot of our testbed is shown in Figure \ref{fig:turtlebot}.

The bulk of computation, including the deep network and the planning, runs on an onboard computer (Nvidia GTX 1060).
The camera is attached to the onboard computer through a USB and supplies the RGB images.
Given an image and the relative goal position, the onboard computer predicts the next waypoint for the robot, plans a spline trajectory to that waypoint, as well as computes the low-level control commands and the corresponding feedback controller. 
The desired speed and angular speed commands are sent to the Kobuki base, which then converts them to PWM signals to execute on the robot.

\subsection{Imperfections in depth estimation} \label{appendix_sec:depth_imperfections}
\begin{figure}[h!]
    \centering
    \includegraphics[width=\textwidth]{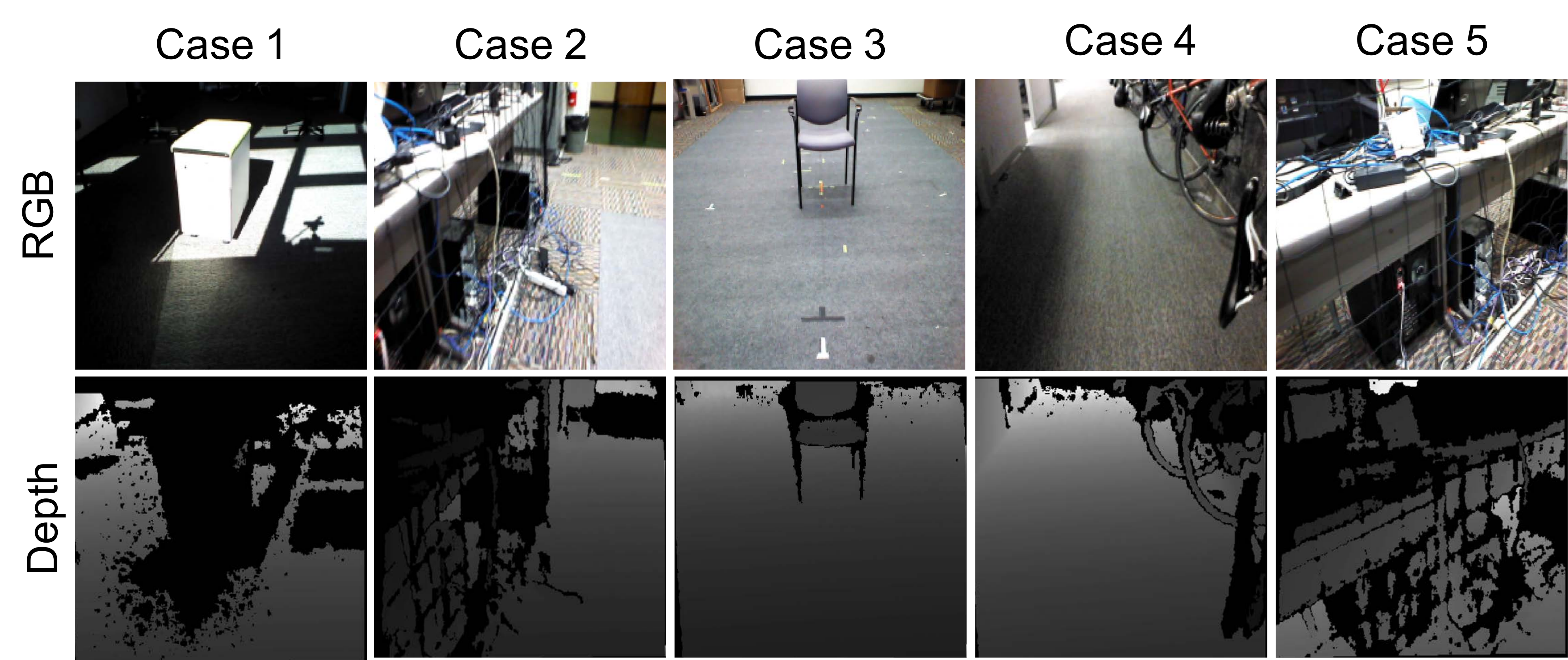}
    \caption{We visualize the RGB images captured by the robot and corresponding depth estimation. The black region corresponds to unknown depth. The depth estimation is inaccurate when the robot encounters shiny, transparent and thin objects, resulting in a significant decline in the performance of a mapping-based approach.}
    \label{fig:depth_estimation}
\end{figure}

In Figure \ref{fig:depth_estimation}, we illustrate some examples of inaccurate depth estimations that we encounter during our experiments.
The black pixels in the depth images correspond to the regions where the depth estimator fails to accurately estimate the depth.
In Case 1, the depth estimation fails because of the presence of sunlight. In Case 2, thin wires and power strips on the floor are not recorded accurately. 
In Case 3, the depth estimation fails due to the shiny, thin legs of the chair.
In Case 4, bike seats and tire frames are not estimated accurately.
In Case 5, shiny, transparent monitors are not recognized by the depth estimator.